%% file: main.tex
\documentclass{article}
\input{preamble}

\title{EgoVine: Post-Training with Branching\\Experience Compiled from Real-World Videos}
\author{
Qinchuan Cheng$^{1}$ \qquad Bowen Lyu$^{2}$ \qquad Zhantao Gong$^{3}$\\
Angela Yao$^{4}$ \qquad Cheston Tan$^{5}$ \qquad Shijie Li$^{6}$\\[3pt]
\small $^{1}$Xi'an Jiaotong University \quad
$^{2}$Beijing Institute of Technology \quad
$^{3}$Nankai University\\
\small $^{4}$National University of Singapore \quad
$^{5}$A*STAR \quad
$^{6}$Institute for Infocomm Research (I2R), A*STAR
}

\iclrfinalcopy

\begin{document}
\maketitle
\lhead{EgoVine preprint}

\begin{abstract}
\input{sections/00_abstract}
\end{abstract}

\input{sections/01_introduction}
\input{sections/02_related_work}
\input{sections/03_problem_setup}
\input{sections/04_branching_experience}
\input{sections/05_post_training}
\FloatBarrier
\input{sections/06_experiments}
\input{sections/07_discussion}
\input{sections/08_conclusion}
\par\vspace{\baselineskip}\noindent\begin{minipage}{\linewidth}
\input{sections/09_statements}
\end{minipage}\par

\clearpage
\begingroup
\raggedright
\bibliographystyle{iclr2027_conference}
\bibliography{references}
\endgroup

\clearpage
\appendix
\input{sections/10_appendices}

\end{document}

%% file: preamble.tex
\usepackage[T1]{fontenc}
\usepackage{iclr2027_conference,times}
\usepackage{amsmath,amssymb,mathtools}
\usepackage{graphicx}
\graphicspath{{figure/}}
\usepackage{xcolor}
\usepackage{booktabs,array,tabularx}
\usepackage{multirow}
\usepackage{algorithm}
\usepackage{algpseudocode}
\usepackage{placeins}
\usepackage{url}
\usepackage{hyperref}
\hypersetup{
  colorlinks=true,
  linkcolor=black,
  citecolor=black,
  urlcolor=blue!45!black,
  pdftitle={EgoVine: Post-Training with Branching Experience Compiled from Real-World Videos},
  pdfauthor={Qinchuan Cheng, Bowen Lyu, Zhantao Gong, Angela Yao, Cheston Tan, Shijie Li}
}

\definecolor{draftblue}{HTML}{365C7D}

\newcolumntype{Y}{>{\raggedright\arraybackslash}X}
\newcolumntype{C}[1]{>{\centering\arraybackslash}p{#1}}
\providecommand{\tightlist}{\setlength{\itemsep}{0pt}\setlength{\parskip}{0pt}}

%% file: sections/00_abstract.tex
Real-world egocentric videos capture multi-step household procedures, but their recorded trajectories leave alternative actions and recovery from the learner's own mistakes largely unexplored. We introduce EgoVine, a post-training framework that turns a fixed demonstration set into interactive experience for long-horizon planning. EgoVine operates in executable symbolic environments compiled from real-world video annotations, where an agent must coordinate object manipulations under partial observation and account for delayed consequences. Restoring a decision state allows the learner to explore alternative actions and compare their eventual outcomes. Union jointly trains action generation and consequence prediction in one model, using the same executions to supervise candidate comparison and recovery. Verified repairs add recovery targets from failed attempts, while comparisons among sampled continuations guide policy updates. The updated planner then returns to the environments, with its own decisions and mistakes shaping the next round of experience. On a shared benchmark, EgoVine-1.5B achieves 81.8\% success on Main planning tasks and 66.7\% on longer task compositions, outperforming all evaluated Flash LLM baselines on both. EgoVine connects real demonstrations to an iterative learning process that trains compact planners on choices and futures beyond the recorded trajectories.

%% file: sections/01_introduction.tex
\section{Introduction}
\label{sec:introduction}

A demonstration shows one way to complete a task. It rarely shows what would have happened after a different action, how to recover from a learner's mistake, or when a seemingly unhelpful step becomes necessary later. To retrieve an object and leave its cabinet closed, for example, an agent may first have to open the cabinet, temporarily violating a condition of the final goal. A learner also reaches states absent from the recording, where several actions may be executable but lead to different futures. Learning requires experience of those choices and their continuations.

Egocentric videos capture rich object interactions and everyday procedures~\citep{hdepic2025}. Their value for planning extends to the objects, state changes, and action requirements that govern other possible procedures. Ego2World~\citep{ego2world2026} compiles these annotations into executable symbolic worlds for planning under partial observation. EgoVine turns these worlds into a source of post-training experience: the learner can revisit a decision, try alternatives under the grounded rules, and follow their consequences beyond the recorded trajectory.

We study \emph{how environments compiled from a fixed set of real demonstrations can turn passive trajectory supervision into interactive learning}. Execution supplies action--consequence targets; restoration makes alternatives comparable from a common decision. Their continuations reveal useful detours, failures, and repairs. These experiences support both supervised learning and policy optimization. As the actor changes, its state visits expose further gaps in the experience available for learning.

EgoVine organizes experience around decisions, their executable alternatives, and what follows each choice. A shared context makes alternatives comparable; a verified suffix can supply a future from a new starting point. This connects three contributions:
\begin{itemize}
\tightlist
\item \textbf{Expanding experience beyond demonstrated paths.} We construct executable alternatives, failures, and repair continuations from demonstrated and policy-visited states.
\item \textbf{Learning aligned lessons from each execution.} Union connects action, consequence, candidate-selection, and bounded-rejoin targets for one planner. Each execution provides corresponding targets for action choice, consequences, and continuation, with the checked outcome determining its supervision.
\item \textbf{Continuing learning through controlled interaction.} BunchFusion organizes restored-state comparisons and compatible reuse; supervision and relative policy updates learn from the resulting branches. The updated actor's state visits then determine which alternatives and repairs to collect next.
\end{itemize}

Our experiments follow this process from available supervision to learning and task performance. Recovery-expanded supervision produces the largest observed training-stage improvement. The compact planners lead complementary benchmark panels; paired traces reveal differences in subgoal choice and prerequisite handling, and fixed-workload measurements quantify compatible reuse.

%% file: sections/02_related_work.tex
\section{Related Work}
\label{sec:related-work}

\paragraph{Grounded worlds as learning environments.}
UniSim learns a visual simulator from diverse static datasets and uses simulated interaction to train policies~\citep{unisim2023}. Ctrl-World develops a controllable video world model for manipulation, using successful imagined trajectories for fine-tuning~\citep{ctrlworld2025}. RoboCasa365 expands household task and scene coverage~\citep{robocasa3652026}, while MagicSim unifies world construction, embodied execution, and trajectory generation in a shared runtime~\citep{magicsim2026}. Ego2World compiles egocentric annotations into symbolic worlds for belief-state planning~\citep{ego2world2026}. EgoVine studies the learning opportunities within a fixed set of these worlds. Restorable states and explicit rules let the learner explore unrecorded decisions, check their consequences, and connect new branches to verified continuations. The research focus is how this experience becomes training supervision and policy credit.

\paragraph{From video observation to interactive learning.}
Seeing-to-Experiencing (S2E) combines video pretraining with reinforcement learning in simulation to acquire reactive navigation while retaining pretrained knowledge~\citep{s2e2025}. EgoVine addresses discrete, object-centered procedures in worlds compiled from the source demonstrations: interaction exposes alternative action sequences and repairs, which supply both supervised targets and policy outcomes. VAGEN learns explicit state estimation and transition reasoning for partially observed visual-agent tasks~\citep{vagen2025}. Union aligns the chosen action, observed effects, candidate preferences, and bounded-rejoin labels around one executed decision. A failed attempt can teach applicability while a checked repair teaches how to continue. The planner learns these correspondences from textual observations and execution feedback.

\paragraph{Comparing intermediate decisions.}
VinePPO estimates credit in mathematical reasoning using Monte Carlo continuations from intermediate prefixes~\citep{vineppo2025}. TreeRL branches at uncertain reasoning steps in mathematics and code~\citep{hou2025treerl}; Tree-GRPO combines shared-prefix sampling with intra-tree and inter-tree advantages for multi-turn question answering~\citep{treegrpo2025}. GiGPO groups recurring states for finer credit in ALFWorld, WebShop, and search-augmented question answering~\citep{gigpo2025}. In EgoVine, intermediate choices change an executable world. Restoring the decision context supports relative policy credit, and executing its alternatives also supplies applicability, effects, selection targets, and checked continuations. Branch comparison thus serves the broader process of expanding and learning from experience across policy-visited states.

\paragraph{Learning from deviations.}
ReAct uses interaction to revise reasoning and plans~\citep{react2023}. AgentRefine constructs feedback-driven correction examples in synthetic environments~\citep{agentrefine2025}; SCoRe learns self-correction from the model's own traces in mathematics and coding~\citep{score2025}. Process supervision labels reasoning steps, while STaR trains on rationales yielding correct answers~\citep{lightman2023lets,zelikman2022star}. DAgger gathers expert actions at learner-visited states to address policy-induced distribution shift~\citep{dagger2011}. EgoVine obtains targets at such states through execution and bounded continuation search. Several valid futures can supply supervision, including a connector to an existing suffix or a new completion. Recovery concerns changes in the world, and each attempt also supplies consequence and selection targets.

%% file: sections/03_problem_setup.tex
\section{Problem Formulation}
\label{sec:problem}

We study goal-directed planning in a partially observed symbolic environment compiled from annotated video. Its state records objects, relations, location, and task phase; transitions are deterministic under fixed rules. The executor maintains the full state, while the actor receives text describing the goal, visible facts, and permitted interaction history.

At each decision, the actor proposes an action. The executor checks its preconditions, applies its effects, and returns an observation and execution feedback. The environment interface updates the actor's information record before the next decision. Opening a visible cabinet, for example, can reveal previously hidden contents. The task is to satisfy the goal within an action horizon $H$; Appendix~\ref{app:formal-setup} gives the formal interaction and objective.

During training, the environment can restore a saved state together with the actor's visible context and remaining horizon. Alternatives can then be compared from the same starting conditions. The full snapshot enables controlled execution; each actor input still contains only its permitted observations.
\FloatBarrier

%% file: sections/04_branching_experience.tex
\section{EgoVine}
\label{sec:post-training}

EgoVine uses execution to extend what can be learned from real demonstrations (Figure~\ref{fig:egovine-framework}). Replaying a demonstration supplies decision states and successful continuations. Exploring from those states, and later from the actor's own state visits, produces alternatives and repairs. Union converts these executions into aligned learning targets; BunchFusion organizes their comparison and reuse. Learning changes the actor, creating new starting points for the next collection.

\input{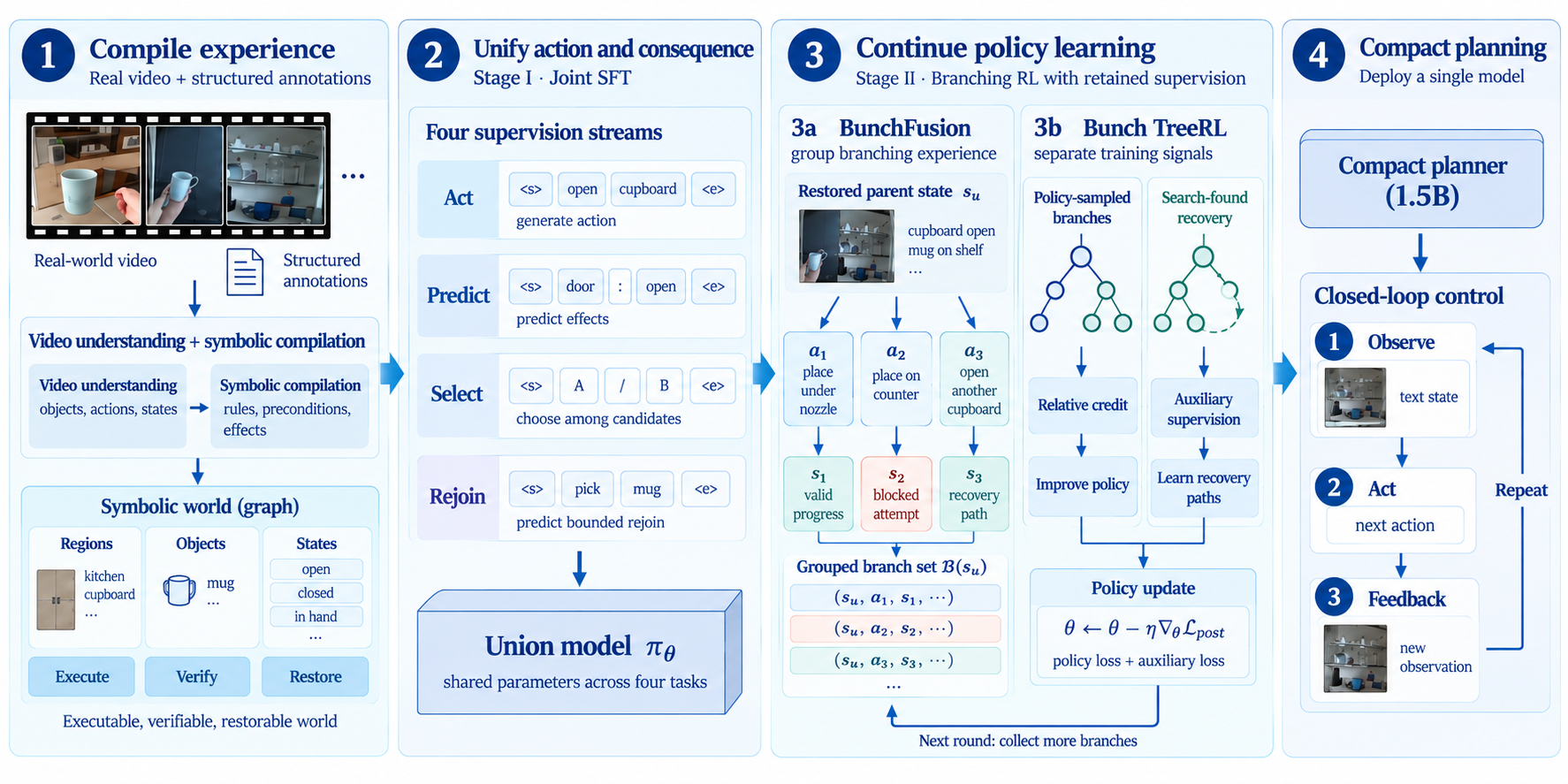}

\subsection{Expanding Experience Beyond the Recorded Trajectory}
\label{sec:experience}

Ego2World~\citep{ego2world2026} compiles annotated objects, relations, actions, and state changes into rules specifying when an action applies and what it changes. EgoVine uses these rules to execute choices that the recording did not show. For each explored decision, it retains the starting context, attempted action, observed effects, and continuation toward the goal. Demonstration replay provides the initial states and checked suffixes; policy visits add states reached through the learner's own choices.

Consider a recorded procedure: open a cabinet, retrieve a bag, close the cabinet, and place the bag at its destination. Restoring the initial state exposes alternatives: early retrieval fails a precondition; opening enables retrieval while temporarily violating the final closed-cabinet condition; a detour creates a new state from which to continue. The resulting experience includes both how actions change the world and how to complete the task after leaving the recorded path. These are the inputs to the aligned supervision and branch comparisons below.

%% file: figure/framework.tex
\begin{figure}[htb]
\centering
\includegraphics[width=\linewidth]{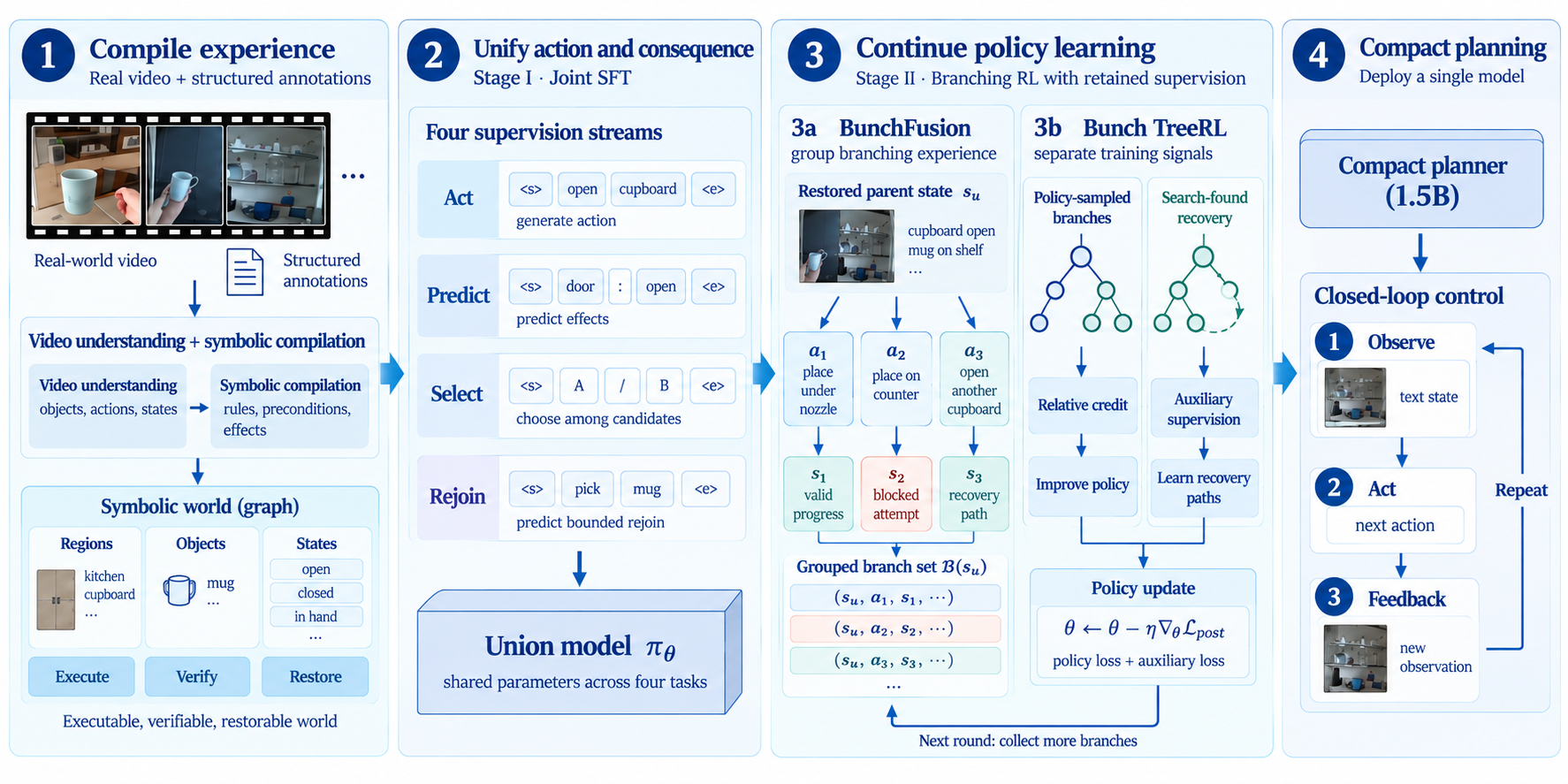}
\caption{EgoVine's executable learning loop. (1) Ego2World compiles video annotations into executable worlds. (2) Union aligns Act, Predict, Select, and bounded-rejoin labels; verified connectors separately supply Act targets. (3a) BunchFusion expands restored states; (3b) the branch-relative update (labeled Bunch TreeRL) and supervision train the actor for further collection. (4) Planning uses textual observations, actions, and feedback.}
\label{fig:egovine-framework}
\end{figure}

%% file: sections/05_post_training.tex
\subsection{Union: Aligned Action--Consequence Learning}
\label{sec:two-stage}

An execution answers several related questions: what can be done, what changes, which alternative is useful, and whether a route back to completion exists. Union turns these answers into Act, Predict, Select, and Rejoin targets for one model (Table~\ref{tab:utm-example}). The targets refer to the same context and action, so a failed attempt supplies consequence supervision even when a different action supplies the positive planning target. The shared training interface preserves this correspondence across tasks.

\begin{table}[H]
\centering
\begin{tabularx}{\linewidth}{@{}lYY@{}}
\toprule
Task & Supplied fields & Supervised target \\
\midrule
Act & Retrieve the bag; cabinet closed & Open the cabinet; valid; cabinet becomes open \\
Predict & Same context and opening action & Applicability, changed state, and goal progress \\
Select & Same context; opening and placing candidates & Opening, since placing requires holding the bag \\
Rejoin & Same context; candidate opening action & Yes: a checked continuation reaches a successful route \\
\bottomrule
\end{tabularx}
\caption{Union views of the same cabinet example. The bag location is known from memory. Rejoin predicts bounded route reachability; verified connector actions supply additional Act targets.}
\label{tab:utm-example}
\end{table}

The joint Act response contains an action followed by its applicability, changed facts, progress, and phase. For example, ``open the cabinet'' is followed by ``valid; cabinet becomes open.'' Both portions train the shared model, with the action generated before its consequence fields. Deployment requests a structured action; the executor applies that action and actual feedback updates the next context and memory. Predicted effects serve as learning targets, rather than observations of an already-known future. Appendix~\ref{app:learning-objectives} specifies the serialization and target masks.

\subsection{BunchFusion: Controlled Branching and Reuse}
\label{sec:bunchfusion}

Restoring a common parent separates the current choice from differences in starting conditions. A frozen actor samples alternatives from the same state, visible context, goal, and remaining horizon. Each receives continuation samples before additional exploration is allocated. Execution records retain applicability, effects, path outcomes, and verified repairs for comparison and supervision.

\paragraph{Following alternatives and reconnecting to known futures.}
Recovery continues toward the goal after a deviation. To find a rejoin, BunchFusion indexes states along replayed training continuations together with their suffixes, searches for a bounded connector from the new state, and replays the connector and suffix to check completion within the remaining horizon. This yields a Rejoin label and verified connector actions for Act supervision. Replanning instead changes the remaining route and can reach a new completion without returning to the original path.

A retained bottle-placement example illustrates why this matters. With the bottle held, clean, and dry, two navigation actions are executable and make no immediate goal progress. One goes directly to the destination and joins a checked seven-action suffix. The other goes to a different surface and needs one navigation action to reconnect. Both routes complete the goal, in eight and nine actions respectively (Appendix~\ref{app:valid-alternatives}). The demonstration's future thus becomes useful from multiple starting points, while continuation cost distinguishes choices that applicability alone cannot rank.

\paragraph{What can be reused.}
Sibling generation can share prompt computation. Restoring or backtracking to a saved state allows another continuation to be explored; repeating a transition under the same state, action, and rules can reuse its checked result. Rejoin reuses a verified suffix after checking it from the reached state. Sharing a sampled policy continuation additionally requires matching visible history, goal, policy, remaining horizon, and decoding constraints. Two paths can reach the same physical state with different actor contexts, so these checks determine which reuse is available. Every path retains its identity and accumulated cost. Appendix~\ref{app:implementation-binding} formalizes these conditions.

\subsection{Learning from Alternative Continuations}
\label{sec:treerl}

\input{figure/bunchfusion_treerl}

Branches provide both outcomes to compare and targets to learn. A failed attempt reveals an unmet precondition; a verified repair provides a continuation the actor did not find. Bunch TreeRL assigns relative credit to sampled choices, while outcome-aware supervision uses the checked actions, effects, and continuations.

\paragraph{Comparing choices through their futures.}
A sampled path is complete when it succeeds or reaches its task horizon, so failed attempts remain part of the comparison. For path $j$, let $S_j$ indicate success, $d_0>0$ and $d_j$ count initially and finally unsatisfied goal predicates and milestones, $I_j$ count inapplicable attempts, and $L_j\le H$ count actions. Its return combines completion, endpoint progress, and execution costs:
\begin{equation}
z_j=S_j+\lambda_d\frac{d_0-d_j}{d_0}-\lambda_i\frac{I_j}{H}-\lambda_s\frac{L_j}{H}.
\label{eq:branch-score}
\end{equation}
The endpoint term evaluates preparation, such as opening a cabinet, through what follows it. Recovery completed by the sampled policy contributes to its return; recovery found by search supplies supervised continuation targets. The coefficients and replay checks are given in Appendix~\ref{app:learning-objectives}.

Let $\mathcal C(u,a)$ contain the comparable completed continuations of action $a$ and $\mathcal A_u$ the alternatives at parent $u$. We average each choice's outcomes, then compare it with its siblings:
\begin{equation}
\widehat Q(u,a)=\frac{1}{|\mathcal C(u,a)|}\sum_{j\in\mathcal C(u,a)}z_j,
\qquad
\widehat A(u,a)=\widehat Q(u,a)-\frac{1}{|\mathcal A_u|}\sum_{a'\in\mathcal A_u}\widehat Q(u,a').
\label{eq:branch-value}
\end{equation}
Each compared choice receives the prescribed continuation coverage. Averaging summarizes the sampled futures of each choice; centering compares them within the same decision. Figure~\ref{fig:bunchfusion-treerl} illustrates this calculation. The update normalizes these advantages and applies a token-level clipped surrogate~\citep{schulman2017ppo}, sharing each branch advantage across its response tokens. Appendix~\ref{app:learning-objectives} gives the full reduction.

\paragraph{Learning from each kind of outcome.}
Examples are sampled by source---demonstration, alternative, failure, or repair---and their checked outcomes determine which fields receive supervision. A record can have several attributes: a repair may also advance the goal or change the plan. Its source determines its sampling group, while these attributes determine its targets.

Demonstrations and successful alternatives supervise Act and Select, and their effects supply Predict targets. Invalid attempts teach applicability and effects, or selection of a checked alternative. Local regression or lack of progress is interpreted through the continuation: an enabling action remains a positive target when its later outcome supports it. Verified repairs teach continuation actions; replanning provides comparisons among feasible routes. Each collection round sets the proportions of these sources, with earlier examples available for retention.

\paragraph{Updating one planner.}
Aligned supervision and relative policy learning act on the same model. They can be scheduled in successive stages or combined in a joint objective. The evaluated checkpoints use supervision followed by a policy update: 0.5B receives Union and recovery-expanded supervision, while 1.5B receives reference-action supervision before branch learning. Appendix~\ref{app:current-recipes} gives these schedules and their active objectives; Appendix~\ref{app:learning-objectives} gives the general formulation.

\subsection{Continued Collection and Learning}
\label{sec:mixture}\label{sec:procedure}

Demonstrations supply initial knowledge; the actor's state visits and mistakes determine what experience to add. Each round freezes the actor for collection, constructs branch targets, and updates the model. Returning to the same worlds with the updated policy exposes different alternatives, failures, and recovery needs. Earlier targets remain available for retention, while policy-dependent continuations are refreshed. This closes the loop from fixed recordings to policy-driven interactive learning (Algorithm~\ref{alg:egovine}). Appendix~\ref{app:refresh} examines a two-stage instance with fixed starting roots and a refreshed collector.

%% file: figure/bunchfusion_treerl.tex
\begin{figure}[t]
\centering
\includegraphics[width=\linewidth]{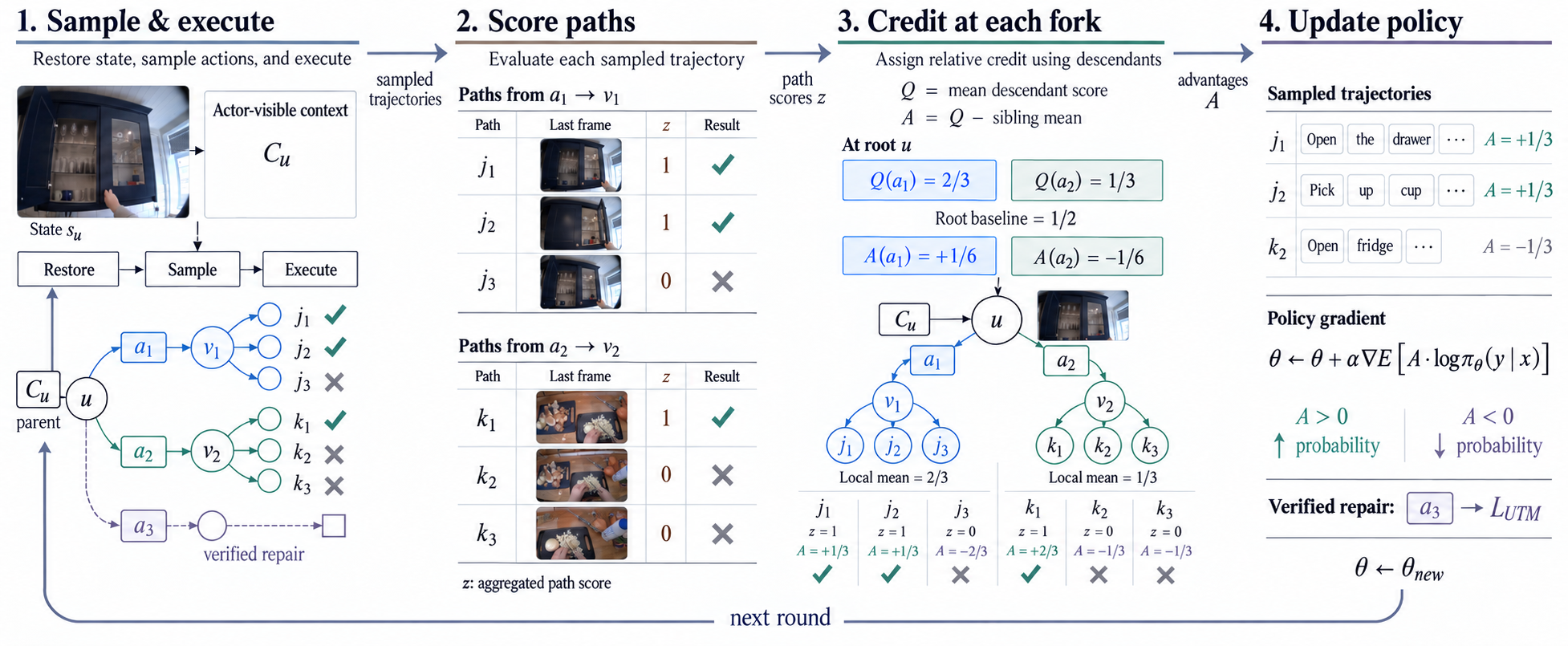}
\caption{Learning from controlled branches. (1) Restore a parent and execute sampled alternatives; search separately verifies a repair. (2) Score the sampled continuations: the illustrated binary outcomes give branch means $2/3$ and $1/3$. (3) Center values at each fork, yielding root advantages $\pm1/6$ and the displayed descendant advantages. (4) Update sampled decisions through policy gradients and use the repair as a Union target ($\mathcal L_{\mathrm{UTM}}$ in the diagram), then collect with the new actor. Advantages are shown before normalization; Appendix~\ref{app:learning-objectives} gives the clipped update and staged or joint schedules.}
\label{fig:bunchfusion-treerl}
\end{figure}

%% file: sections/06_experiments.tex
\section{Experiments}
\label{sec:experiments}

We examine three links in the learning process: what supervision execution reveals beyond a recorded trajectory, how the actor changes after learning from expanded experience, and how the resulting planners perform across task types.

\subsection{What Supervision Does Branching Expose?}
\label{sec:experience-evidence}

A successful demonstration supplies one completion, but an unsuccessful actor episode can also pass through states with useful continuations. Bounded search from 487 SFT-policy contexts finds 401 replay-verified repairs, including 200 from episodes that ultimately failed. Retaining only successful rollouts would omit these repair targets. The contexts come from actor trajectories in training worlds, and replay checks each witness from its saved state (Appendix~\ref{app:evidence}).

Branching also exposes multiple feasible futures at a single decision. The retained bottle-placement example in Section~\ref{sec:bunchfusion} produces eight- and nine-action completion routes from two executable choices with identical immediate progress. Each route supplies consequence and continuation targets, and their different costs make a selection comparison possible. Together, these records demonstrate how the same grounded world yields supervision for choices and states beyond its recorded procedure.

\subsection{How Does the Actor Change with Expanded Experience?}
\label{sec:learning-evidence}

We follow the current 0.5B checkpoint through Union initialization, recovery-expanded supervised training, and its subsequent policy update. All stages are evaluated on the same 169 cases from 61 tasks with two fixed evaluation seeds, holding observations, decoding, and the decision allowance constant. Appendix~\ref{app:final-training} gives the shared settings and the 338 episodes per stage.

Success rises from 215/338 (63.6\%) after Union to 269/338 (79.6\%) after recovery-expanded supervision and 274/338 (81.1\%) for the released checkpoint. The clear step change occurs when the original Union targets are supplemented with recovery experience; the final policy stage is retained in the released checkpoint and evaluated under the same fixed panel.

The 1.5B parent comparison and the fixed-experience controls are retained in Appendix~\ref{app:final-training} for checkpoint provenance and reproducibility. Across the staged results, the clearest measured improvement comes from adding recovery experience to the supervised pool; the later policy stage defines the released checkpoint without being treated as a separate benchmark claim.

\subsection{Planning Performance and the Decisions Behind It}
\label{sec:primary-comparison}

\paragraph{Benchmark and models.}
\label{sec:splits}
We compare EgoVine-0.5B and 1.5B, initialized from Qwen2.5-Instruct~\citep{qwen252024}, with Qwen Flash, DeepSeek Flash, and Luna on the same 99 Main tasks, 51 Long tasks, 100 Recovery cases, and 100 Deviation cases. Each panel uses a common executable-action protocol for all models: the same tasks, permitted observations, parser, deployment budget, success criterion, and 12-decision allowance. The actor observes the goal, visible state, and feedback as text. Appendix~\ref{app:final-benchmark} gives the exact execution protocol and training recipes.

\begin{figure}[tb]
\centering
\includegraphics[width=\linewidth]{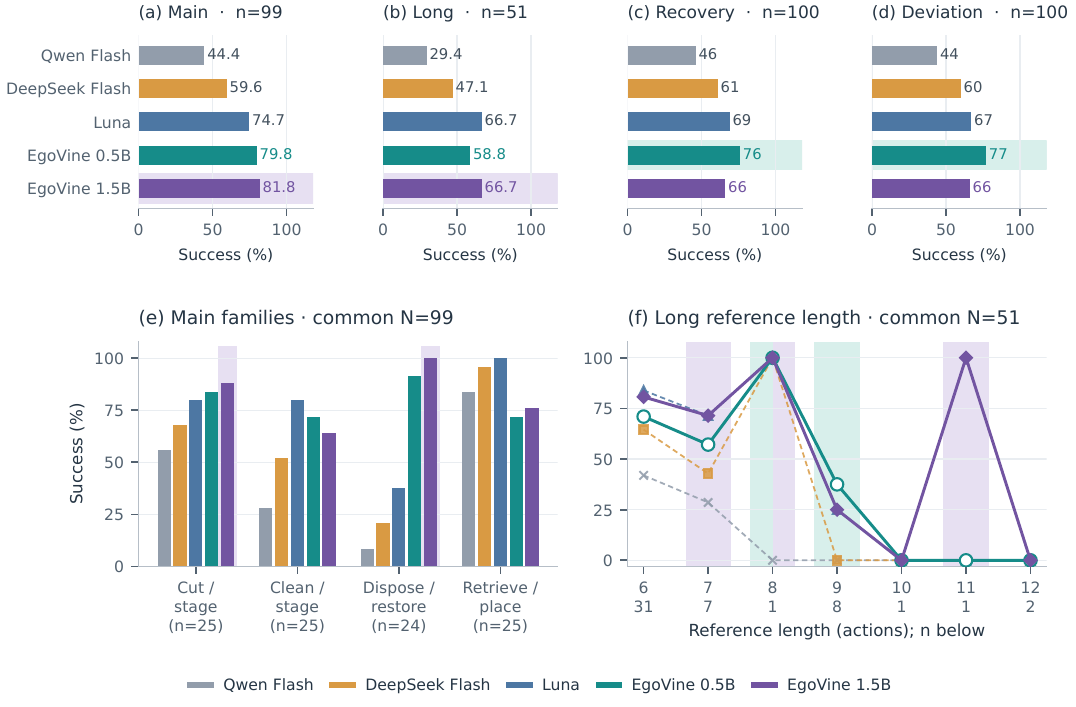}
\caption{Five-model benchmark on identical tasks within each panel. (a--d) Overall success. (e) Main-family success. (f) Long success by executable reference length on the common 51-task set, with group counts below each length. Pale teal and purple mark positive best or tied-best observed values for EgoVine-0.5B and 1.5B. Lines connect discrete reference-length groups; counts indicate the support for each point.}
\label{fig:benchmark-overview}
\end{figure}

Table~\ref{tab:benchmark-system-r0} and Figure~\ref{fig:benchmark-overview} show complementary strengths. On Main, 1.5B achieves 81.8\% success, 7.1 percentage points above Luna and 22.2 above DeepSeek Flash; 0.5B reaches 79.8\%. On Long, 1.5B and Luna tie at 66.7\%, followed by 0.5B at 58.8\%. On Recovery and Deviation, 0.5B has the highest success at 76\% and 77\%, exceeding Luna by seven and ten points; 1.5B reaches 66\% on both. These rankings describe the observed scores; paired intervals for all EgoVine--external comparisons are in Appendix~\ref{app:benchmark-uncertainty}.

\input{tables/benchmark_system_r0}

\paragraph{Which procedures distinguish the models?}
On cutting/staging and disposal/restoration, 1.5B achieves 88\% and 100\%, compared with Luna's 80\% and 37.5\%; 0.5B reaches 84\% and 91.7\%. Luna leads cleaning and retrieval at 80\% and 100\%, versus 1.5B's 64\% and 76\%. Long reference lengths span six to twelve actions: 31 tasks have six-action references, while each later group contains one to eight tasks. The family and discrete-length profiles locate the procedures contributing to each planner's aggregate result.

\paragraph{Subgoal choice in extended planning.}
\label{sec:evidence-analysis}
For behavior analysis, we evaluate both fixed EgoVine checkpoints on the complete paired panels of 100 instances each (Appendix~\ref{app:v2-behavior}). On Long, 1.5B completes 73 tasks versus 56 for 0.5B, a 17-point difference with a task-cluster 95\% interval of $[6,28]$. Yet it has more non-executable decisions: 11.7\% versus 7.5\%. Local applicability therefore explains only part of the planning profile.

In a paired carrot-and-board task, both models cut and place the carrot. The 1.5B model then washes, dries, and stages the required board, completing the task in nine decisions. The 0.5B model instead washes the sink and moves the carrot again; its actions remain executable, but the board subgoal is unfinished at the cap. What matters is which future an applicable action enables. Preparation and temporary regression also occur in failed trajectories, so their usefulness is evaluated through the ensuing completion, as in the branch comparisons (Appendix~\ref{app:v2-preparation-analysis}).

\paragraph{Preserving prerequisites during recovery.}
Recovery and Deviation favor 0.5B by ten and eleven points on the paired panels, with task-cluster intervals $[-2,21]$ and $[0,22]$. On Recovery, it makes 9.4\% non-executable decisions versus 24.7\% for 1.5B. Among parsed actions, inapplicability remains lower at 9.4\% versus 21.2\%, showing that prerequisite failures accompany the completion difference alongside output-format errors. In a saved case, 0.5B retrieves an object before closing its container and completes placement in four decisions. The 1.5B model closes first and attempts placement without the object, later retrieving it but exhausting the allowance. Immediate goal progress can obstruct the remaining procedure (Appendix~\ref{app:v2-cases}).

\paragraph{Completion within the decision budget.}
\label{sec:quality-cost}
The stronger checkpoint in each setting also completes more cases early. On paired Long tasks, 1.5B completes 54 by decision eight versus 38 for 0.5B, and 69 versus 49 by decision ten. Mean decisions over all 100 tasks, including failures, are 8.54 and 9.47. Recovery reverses the pattern: 0.5B completes 52 cases by decision four versus 34, with means of 6.67 and 7.87 (Figure~\ref{fig:v2-budget}).

\begin{samepage}
\paragraph{Reusing computation during experience construction.}
On a fixed 569-candidate workload, snapshot restoration and transition caching reduce CPU execution time from 23.820 to 6.720 seconds ($3.54\times$), with physical transitions falling from 2,551 to 524. For 248 retained requests from the actual 1.5B collector, prompt-KV reuse reduces fixed-token scoring time from 141.138 to 132.542 seconds ($1.065\times$), preserving token log probabilities. These measurements quantify reuse within the environment and model computations; Appendix~\ref{app:reuse-measurements} gives the workloads, generation measurements, and memory costs.
\par\end{samepage}

%% file: tables/benchmark_system_r0.tex
\begin{table}[H]
\centering
\setlength{\tabcolsep}{4.5pt}
\begin{tabularx}{\linewidth}{@{}Yrrrr@{}}
\toprule
Model & Main & Long & Recovery & Deviation \\
\midrule
Qwen Flash & 44.4 & 29.4 & 46.0 & 44.0 \\
DeepSeek Flash & 59.6 & 47.1 & 61.0 & 60.0 \\
Luna & 74.7 & \textbf{66.7} & 69.0 & 67.0 \\
EgoVine 0.5B & 79.8 & 58.8 & \textbf{76.0} & \textbf{77.0} \\
EgoVine 1.5B & \textbf{81.8} & \textbf{66.7} & 66.0 & 66.0 \\
\midrule
Tasks / cases & 99 & 51 & 100 & 100 \\
\bottomrule
\end{tabularx}
\caption{Success (\%) on the same tasks for all five models under the shared executable-action protocol. Best values, including ties, are bold. All panels allow 12 decisions; the exact execution and decoding rules are given in Appendix~\ref{app:final-benchmark}.}
\label{tab:benchmark-system-r0}
\end{table}

%% file: sections/07_discussion.tex
\section{Discussion and Limitations}
\label{sec:discussion}

Executable worlds let the learner return to a demonstration with new questions: which alternatives remain useful, what happens after a mistake, and how a known future can be reached again. The verified repairs and recovery-expanded training results support this role for interactive experience. Union supplies aligned targets for action, consequence, comparison, and recovery; the released checkpoints are evaluated by completed task behavior, while the supplementary probes characterize the supervision interface rather than serving as standalone capability claims.

The results concern video-grounded symbolic worlds, including sources used during training and development. Rule accuracy and coverage determine which alternatives can be explored. Applying the process elsewhere requires execution, restoration, and verification; transfer to new video sources remains to be measured. The reuse experiments establish savings in environment transitions and model computation on fixed workloads. Full-cycle training costs and the value of suffix reuse depend on the compatible branches encountered during collection.

%% file: sections/08_conclusion.tex
\section{Conclusion}
\label{sec:conclusion}

EgoVine gives real demonstrations a continuing role in learning beyond imitation of their recorded paths. Their executable worlds let a planner explore alternatives, check consequences, and discover ways to continue after its own mistakes. Aligned supervision and relative branch comparisons turn that experience into learning signals, while the updated actor guides further collection. Verified repairs, gains from recovery-expanded supervision, and measured computational reuse demonstrate complementary parts of this process: a fixed demonstration can support learning about decisions and futures it never explicitly recorded.

%% file: sections/09_statements.tex
\section*{Reproducibility Statement}

We will release the data, compiled tasks and rules, code, weights, configurations, and execution records with source and checkpoint identities. Appendices~\ref{app:data}--\ref{app:evaluation} describe the procedures and benchmark configuration.

\section*{Ethics Statement}

The source data permit the intended use and release. Distribution will preserve attribution and applicable access conditions. Egocentric recordings may reveal people and private environments, so released examples must respect the source privacy requirements. Physical deployment would require independent validation and execution safeguards.

\section*{AI Use Statement}

Language models assisted with manuscript organization, editing, and drafting. The authors are responsible for the final text, references, implementation, and empirical claims. Release documentation will distinguish model-generated research supervision from editorial assistance.

%% file: sections/10_appendices.tex
\section{Sources, Rules, and Executable Experience}
\label{app:data}

\paragraph{Source and rule provenance.}
Ego2World supplies the executable worlds compiled from HD-EPIC annotations. Demonstrations, tasks, rules, branches, repairs, and prompts retain their source video and construction procedure. The present evaluation measures planning in these grounded worlds, including sources used during training and development. Task identities distinguish a new decision or goal composition from a new source video.

\paragraph{Execution records.}
A decision record contains the saved state, visible context, goal, attempted action, successor, collection checkpoint, and remaining horizon. Path identifiers preserve the origin and multiplicity of repeated draws. Certificates add continuation coverage, progress labels, repair costs, and rejoin-state and suffix identifiers. The actor receives the permitted observation projection; the executor uses its full state for transition checks. Shared suffixes retain their original identities so that repeated references to the same execution are counted consistently.

\subsection{Formal Interaction and Task Objective}
\label{app:formal-setup}

The symbolic environment $\mathcal E$ has deterministic transition operator $T_{\mathcal E}$ under its fixed rules. Its full state is $s_t$, goal is $g$, and action horizon is $H$. The actor receives observations through the projection $\Omega_{\mathcal E}$.

Let $\beta_t$ be a symbolic record of available facts and $h_t$ the retained interaction history. Together they give context $c_t=F(g,\beta_t,h_t)$. A decision is
\begin{equation}
a_t\sim\pi_\theta(\cdot\mid c_t),\qquad
(s_{t+1},e_t)=T_{\mathcal E}(s_t,a_t),\qquad
\beta_{t+1}=U\!\left(\beta_t,a_t,\Omega_{\mathcal E}(s_{t+1},e_t)\right).
\label{eq:transition}
\end{equation}
The executor checks applicability, effects, and goal progress; only the permitted observation $\Omega_{\mathcal E}$ reaches the actor. The environment interface applies $U$ to update the information record from this feedback. For example, opening a visible cabinet can reveal previously hidden contents. The objective is completion within horizon $H$:
\begin{equation}
J(\theta)=\mathbb E_{(\mathcal E,g)\sim\mathcal D,\,\tau\sim\pi_\theta}
\left[\mathbf 1\{\exists\,t\le H:\ g(s_t)=1\}\right].
\label{eq:task-objective}
\end{equation}

\subsection{Execution Certificates}
\label{app:certificate-semantics}

Fix the rule version, parser, goal semantics, and a saved initial state. A transition check parses the proposed action, evaluates its preconditions, and recomputes the successor under those rules. A positive applicability certificate establishes that the action passes the compiled preconditions in that state. Continuation evidence then evaluates its usefulness for reaching the goal.

For a repair sequence $w=(a_0,\ldots,a_{L-1})$, the checker replays every action from the saved state, checks applicability at each step, and verifies the endpoint condition and budget. Assuming a deterministic, correctly implemented executor, acceptance establishes the existence of this executable path in the compiled world, by induction over the replayed transitions. A rejoin additionally checks the suffix from the reached state under the same goal, phase, and remaining budget. Reuse of a policy continuation also satisfies the actor-context conditions in Eq.~\ref{eq:fusion-key}.

Bounded search records a verified witness when one is found and leaves unsuccessful searches unresolved unless a separate environment proof establishes infeasibility. A witness certifies an executable route; policy evaluation measures how reliably the learned actor completes it.

\noindent\begin{minipage}{\linewidth}
\paragraph{The path-score bound.}
For completed paths, goal achievement implies $d_j=0$, while $0\le I_j\le L_j\le H$. With $(\lambda_d,\lambda_i,\lambda_s)=(0.25,0.10,0.02)$ in Eq.~\ref{eq:branch-score}, a successful path scores at least $1+0.25-0.10-0.02=1.13$. An unsuccessful path scores at most $0.25$, since $d_j\ge0$ and both cost terms are nonpositive. The resulting gap is at least $0.88$ for completed paths. This bound concerns realized path scores; expected action values also depend on the distribution of continuations and their costs. Outcomes are taken from actor paths observed through completion or their task horizon. Interrupted collections remain incomplete, and searched repairs supply auxiliary targets.
\end{minipage}

\section{Training Procedure and Implementation}
\label{app:training}

\input{algorithms/egovine_training}

\paragraph{Union supervision.}
Stage I combines demonstration replay with verified candidate-comparison and repair examples from training worlds. One model learns all four task families in Figure~\ref{fig:egovine-framework}, using the output formats associated with each target. Stage II adds policy-generated alternatives and their execution evidence. Examples retain their origin as demonstrations, sampled branches, or search-assisted targets.

\begin{table}[H]
\caption{Sources of Union supervision. Search-assisted targets are auxiliary evidence unless they also satisfy the requirements for current-policy credit.}
\label{tab:utm-tasks}
\centering\small
\setlength{\tabcolsep}{4pt}
\begin{tabularx}{\linewidth}{@{}>{\raggedright\arraybackslash}p{2.9cm}Y Y@{}}
\toprule
Task family & Stage I & Stage II \\
\midrule
Action generation & Demonstrated actions under visible contexts. & Eligible branch actions and retained demonstrations. \\
Consequence prediction & Replay evidence for supplied actions. & Applicability and relevant effects of sampled candidates. \\
Branch selection & Verified candidate comparisons from training worlds. & Sibling outcomes with comparable evidence. \\
Recovery/rejoining & Demonstrated or executor-verified repairs and suffixes. & Verified repairs and rejoins from collected states. \\
\bottomrule
\end{tabularx}
\end{table}

\paragraph{Collection and update.}
A round begins with a frozen collection checkpoint. Parent states are visited in training worlds, sibling actions are sampled, and each is executed from a restored snapshot. Comparable continuation coverage precedes any additional adaptive allocation. Compatibility checks follow Eq.~\ref{eq:fusion-key}. The update uses only eligible policy samples for relative credit; searched witnesses and replay enter their respective auxiliary objectives. Under the mixture sampler, a role's expected contribution depends on both its sampling proportion $\alpha_k$ and loss weight $\lambda_k$. Empty role pools are excluded and the active proportions are normalized.

The training protocol specifies the starting model, output format, token masks, optimizer, sampling schedule, and seeds. Policy ratios use the probabilities recorded during collection on the same response support used for the update. These choices determine the concrete realization of the objective in Eq.~\ref{eq:post-loss}.

\paragraph{Initialization and retention.}
Appendix~\ref{app:final-training} follows each current checkpoint from its exact supervised parent. The earlier 0.5B controls in Appendix~\ref{app:completed-settings} use the same Union starting point, before the current recovery-expanded stage. The Qwen2.5 models~\citep{qwen252024} share the semantic task interface; each size retains its own training and evaluation comparisons. Retention examples come from training sources and preserve earlier action--consequence targets as new experience is added.

\input{sections/19_learning_objectives}

\input{sections/12_implementation_details}

\FloatBarrier
\input{sections/14_final_benchmark_details}
\FloatBarrier
\input{sections/17_v2_behavior}
\FloatBarrier
\section{Retained Evidence and Qualitative Cases}
\label{app:results}
\label{app:evidence}

The retained training records below illustrate aligned supervision, alternative continuations, and bounded repair construction in the 0.5B development study.

\paragraph{From deviations to repair targets.}
The broad recovery pool contains 1,399 contexts spanning 47 tasks and 10 training videos: 373 unknown-entity feedback contexts, 373 known-entity precondition failures, and 653 valid world deviations. Independent replay verifies a completion witness for 1,330 contexts; 69 remain bounded-search misses. These reference-assisted witnesses contain 6,832 action occurrences and 4,073 unique prompt--response pairs.

A separate pool mined from SFT trajectories on training videos contains 487 contexts: 401 have replay-verified repair paths and 86 remain unresolved within budget. Of the verified contexts, 200 come from episodes the actor ultimately failed. The counts describe intermediate contexts, and trigger categories can overlap. Failed searches and successful witnesses are both retained. This inventory measures available repair supervision; the current checkpoints' recovery behavior is evaluated in Appendix~\ref{app:v2-behavior}, and their within-parent training comparisons appear in Appendix~\ref{app:final-training}.

\paragraph{A necessary temporary regression.}
In the cabinet task, opening the cabinet reveals the bag but violates the final closed-cabinet predicate. The bottle has already been washed and staged. A separately replayed witness retrieves the bag, closes the cabinet, navigates to the microwave surface, and places the bag there. Retrieval leaves the remaining-goal count unchanged; closing reduces it, navigation leaves it unchanged again, and placement completes the goal. This sequence demonstrates useful preparation beyond immediate predicate progress. The source actor episode also eventually succeeded.

{\raggedright\noindent Source: video \path{P01-20240202-171220}, composite task \path{6fc5f611200e1442}.\par}

\paragraph{Two executable alternatives with different continuation costs.}
\label{app:valid-alternatives}
The retained Union records for the same task at reference-prefix index 3 compare navigation to \texttt{microwave\_02} and \texttt{microwave\_03}. The bottle is held, clean, and dry; its destination is \texttt{microwave\_03}. Both records share identical actor context and are valid with no immediate progress. Both carry a positive bounded-rejoin label. Navigation to \texttt{microwave\_03} reaches the indexed suffix directly; navigation to \texttt{microwave\_02} requires one connector action, navigation to \texttt{microwave\_03}. The shared seven-action suffix places the bottle, navigates to the bag, opens the cupboard, retrieves the bag, closes the cupboard, navigates to the destination, and places the bag. Candidate plus connector plus suffix therefore costs eight or nine actions. Deterministic replay of both routes completes the formal goal. Continuation cost distinguishes two feasible completions even when immediate progress is tied.

The source records end in \texttt{::3::wm::1} and \texttt{::3::wm::9} in the retained Union training pool. The evidence directory \path{qualitative_evidence/valid_alternatives} includes the original records, task, shared-context hash, and complete replay traces. The local check uses connector depth one and a candidate-pool limit of 64; it verifies existing labels without model calls or changes to benchmark counts. This context is a replayed training prefix, not an additional measured policy rollout.

\paragraph{A repair after an unsuccessful policy trajectory.}
A second state is saved after a location-precondition error in a failed actor episode. The state admits a verified sequence that washes and dries a board, navigates to its destination, and places it there, supplying a continuation target from a recoverable intermediate state.

{\raggedright\noindent Source: video \path{P01-20240202-195538}, composite task \path{1b8237499db6e3fb}.\par}

\paragraph{Artifact provenance.}
The repair examples and policy inventory are retained in \path{actual_policy_recovery_evidence_v1.tar.gz}; the broad-pool manifest accompanies \path{broad_recovery_teacher_v1_retention.tar.gz}. Records link checkpoints and episodes to saved states, search budgets, witness actions, and transition checks. Repeated targets preserve their shared origin.

\FloatBarrier
\input{sections/18_current_training_controls}
\FloatBarrier
\input{sections/13_completed_experiments}

\FloatBarrier
\input{sections/20_supplementary_experiments}

%% file: algorithms/egovine_training.tex
\begin{algorithm}[H]
\caption{EgoVine: expand executable experience, update the shared actor, and collect again.}
\label{alg:egovine}
\small
\begin{algorithmic}[1]
\Require Training worlds and demonstrations, model $\theta$, collection budget $\mathcal B$, supervision mixture $\alpha$
\Ensure Compact iterative planner $\theta$
\State Replay training demonstrations to obtain decision states, aligned targets, and suffix anchors.
\State Execute alternatives and verify bounded repairs in training worlds to construct Union targets.
\State Initialize $\theta$ with the configured supervised targets (Eq.~\ref{eq:utm-loss}).
\For{each post-training round}
  \State Freeze the collection and reference policies at the round's starting model.
  \State Visit training worlds with the actor; save decision states, visible contexts, and remaining horizons.
  \State Sample sibling actions from a common parent, restoring it before each execution.
  \State Continue each sibling with comparable coverage within $\mathcal B$.
  \State Share compatible computation while preserving each path and its cost.
  \State Search from deviations to suffix anchors or new completions; verify the resulting paths.
  \State Retain completed sampled outcomes for relative credit.
  \State Compute sibling values and normalized advantages using Eqs.~\ref{eq:branch-value} and~\ref{eq:adv-normalization}.
  \State Construct Act, Predict, Select, and Rejoin targets from checked branch and repair evidence.
  \State Update $\theta$ with the configured supervision and policy schedule (Eq.~\ref{eq:post-loss}).
  \State Keep the checkpoint prescribed by the training schedule; refresh policy-dependent continuations.
\EndFor
\State \Return $\theta$ for closed-loop planning.
\end{algorithmic}
\end{algorithm}

%% file: sections/19_learning_objectives.tex
\subsection{Target Serialization and Learning Objectives}
\label{app:learning-objectives}

\paragraph{Union targets.}
For context $c_t$, the joint action--consequence response is
\begin{equation}
y_t=\bigl(\langle\mathrm{ACT}\rangle,a_t,\langle\mathrm{EFFECT}\rangle,v_t,\delta_t,p_t,q_t,\langle\mathrm{END}\rangle\bigr),
\label{eq:utm-format}
\end{equation}
where $v_t,\delta_t,p_t,q_t$ denote applicability, effects, progress, and phase. The action precedes its consequence fields. Predict supplies the action in the input and supervises its consequences; Select supplies candidates and supervises the checked preference; Rejoin supplies a candidate and supervises bounded route reachability. Verified connector actions form separate Act targets. For the selected task and target positions $m_i$, the supervised loss is
\begin{equation}
\mathcal L_{\mathrm{Union}}=-\mathbb E_{(c,y)\sim\mathcal D_{\mathrm{sup}}}\sum_i m_i\log p_\theta(y_i\mid c,y_{<i}).
\label{eq:utm-loss}
\end{equation}
Prompt positions carry no loss. The configured supervised stage selects the available task families; Appendix~\ref{app:current-recipes} specifies this selection for the evaluated models.

\paragraph{Path returns and comparable continuations.}
Equation~\ref{eq:branch-score} uses $(\lambda_d,\lambda_i,\lambda_s)=(0.25,0.10,0.02)$. Distance counts unsatisfied goal predicates and milestones. A completed sampled path ends in success or reaches its allotted horizon; its endpoint supplies $d_j$. Independent replay checks the path before its return enters a comparison.

To make the availability requirement explicit, let $m_j^{\mathrm{cmp}}$ indicate whether continuation $j$ is completed, verified, and comparable under the specified starting context and coverage. The branch value is
\begin{equation}
\widehat Q(u,a)=\frac{\sum_{j\in\mathcal C_{\mathrm{all}}(u,a)}m_j^{\mathrm{cmp}}z_j}{\sum_{j\in\mathcal C_{\mathrm{all}}(u,a)}m_j^{\mathrm{cmp}}}.
\label{eq:branch-value-masked}
\end{equation}
The denominator must be nonzero and every compared sibling must meet its prescribed coverage. The set $\mathcal C(u,a)$ in Eq.~\ref{eq:branch-value} contains these eligible continuations, preserving their sample multiplicities. The concrete six-slot collector excludes incomplete groups rather than changing sibling coverage after observing outcomes. Branch-relative advantages are normalized within each fork:
\begin{equation}
\widetilde A(u,a)=\frac{\widehat A(u,a)}{\operatorname{sd}_{a'\in\mathcal A_u}[\widehat Q(u,a')]+\tau},
\qquad \tau=10^{-4},
\label{eq:adv-normalization}
\end{equation}
using the population standard deviation. The implementation below specifies terminal-prefix slots and zero-variance groups.

\paragraph{Outcome-aware mixture.}
For source groups $\mathcal D_k$, sampling proportions $\alpha_k$, target masks $m_k$, and target cross-entropy $\ell_k$,
\begin{equation}
\mathcal L_{\mathrm{mix}}=\sum_k\alpha_k\lambda_k\mathbb E_{z\sim\mathcal D_k}[m_k(z)\ell_k(z)],\qquad\sum_k\alpha_k=1.
\label{eq:mixture}
\end{equation}
A primary source role determines the sampling group; the checked outcome determines the applicable target fields. A repair can also be a successful alternative, without being counted as a second independent record. Each round fixes $\alpha_k$ and loss weights $\lambda_k$. Empty groups are excluded before renormalization. Retained examples preserve their earlier targets, and any intentional use in both mixture and retention terms preserves the intended total weight.

\paragraph{Policy and joint objectives.}
Let $\rho_i$ be the ratio between current and recorded behavior probabilities at an eligible response token, and let $N_{u,a}$ count the response tokens in the corresponding sampled segment. With normalized branch advantage $\widetilde A=\widetilde A(u,a)$, the token-level clipped surrogate is
\begin{equation}
\mathcal L_{\mathrm{tree}}=-\mathbb E_{u,a}\left[\frac{1}{N_{u,a}}\sum_i\min\left(\rho_i\widetilde A,\operatorname{clip}(\rho_i,1-\epsilon,1+\epsilon)\widetilde A\right)\right].
\label{eq:tree-surrogate}
\end{equation}
The reference checkpoint remains fixed during an update round. Supervision and policy learning fit the general objective
\begin{equation}
\mathcal L_{\mathrm{post}}=\mathcal L_{\mathrm{tree}}+\lambda_{\mathrm{mix}}\mathcal L_{\mathrm{mix}}+\lambda_{\mathrm{Union}}\mathcal L_{\mathrm{Union}}^{\mathrm{ret}}+\lambda_{\mathrm{KL}}\mathcal R_{\mathrm{KL}}.
\label{eq:post-loss}
\end{equation}
A schedule can use the supervised objectives before policy learning or activate them together. Both evaluated checkpoints use successive stages and set auxiliary replay weights to zero in the final policy stage. The recipes in Appendix~\ref{app:current-recipes} identify the actual supervision, response support, and coefficients; the following section gives the full-response policy reduction.

%% file: sections/12_implementation_details.tex
\subsection{Outcome-Based Response Update and Reuse Details}
\label{app:implementation-binding}

This section specifies the full-response policy update used by the current 1.5B checkpoint. It gives the policy reduction explicitly and describes how aligned supervision enters the general joint objective. Appendix~\ref{app:current-recipes} identifies the supervision and policy stages actually run for each benchmark checkpoint; the 0.5B policy stage uses conditional argument-fork updates.

\paragraph{Branch and certificate notation.}
At a restored parent, sibling actions are sampled as
\begin{equation}
a_u^{(1)},\ldots,a_u^{(B)}\sim\pi_{\theta_{\mathrm{old}}}(\cdot\mid c_u).
\label{eq:sibling-sampling}
\end{equation}
Their certificates collect applicability $v$, effects $\delta$, progress $p$, repair witness $w^{\mathrm{rep}}$, rejoin information $\jmath$, continuation outcomes $\ell$, and availability masks $m$:
\begin{equation}
\kappa_{u,a}=(v,\delta,p,w^{\mathrm{rep}},\jmath,\ell,m).
\label{eq:certificate}
\end{equation}

\paragraph{Compatibility and reuse.}
A saved parent contains the world state, actor information and history, goal, phase, rule version, and remaining horizon. Reusing an environment transition requires the same transition-relevant state, action, and rules. Policy continuations additionally require
\begin{equation}
K(u)=\bigl(K_{\mathrm{env}}(s_u),c_u,g,\theta_{\mathrm{old}},H_u,\chi_u\bigr),
\label{eq:fusion-key}
\end{equation}
where $K_{\mathrm{env}}$ includes task phase and rule version, and $\chi_u$ contains decoding and continuation constraints. Policy reuse matches the visible history as well as the physical state. Paths retain their sample identities and multiplicities, and shared computation is counted once. A cache hit refers to the original draw and observation. Filtered proposals enter policy credit with their actual behavior distribution and sampling correction; otherwise they supply auxiliary targets. Additional adaptive allocation requires its own treatment of selection bias beyond the reserved sibling coverage.

\paragraph{Objective conventions.}
For a nonempty source pool $\mathcal D_k$, sampling follows $q_\alpha(z)=\alpha_{r(z)}/|\mathcal D_{r(z)}|$, with proportions and target masks defined in Appendix~\ref{app:learning-objectives}. The policy objective averages clipped response-token ratios weighted by branch advantages. Its optimization unit is the response token, with the branch supplying the outcome signal. This surrogate differs from whole-action importance weighting and makes no unbiasedness claim for Eq.~\ref{eq:task-objective}.

\paragraph{Collection and eligibility.}
The collector samples two actor prefixes, each of up to three actions, from the same saved root. Unless a prefix terminates, its state and visible context are cloned to sample three independent continuations through the remaining action horizon. The resulting six weighted leaves share their sampled prefixes. Decoding uses temperature one, top-$p$ one, and top-$k$ zero; every generated response token retains its behavior log probability. All sampled branches are retained with their outcomes. Reference actions and searched repairs enter the supervised channels separately.

An already-terminal prefix is represented by three virtual descendant slots with the same observed outcome and no new actions. These slots preserve equal root-branch weights without creating a suffix fork or extra model calls. A complete leaf contains its prefix and suffix, each linked to the original physical decisions. A task-horizon stop is a completed path with its observed success label. Policy updates use complete, independently replayed six-slot groups; interruptions and missing records exclude the whole comparison.

\paragraph{Reward and evidence channels.}
The reward checker verifies state hashes, the goal-distance chain, applicability, and the terminal goal predicate before evaluating Eq.~\ref{eq:branch-score}. An inapplicable action must leave the state and distance unchanged. Distance reduction is computed once from the initial and final distances, rather than summing positive progress events. Table~\ref{tab:evidence-routing} separates this policy evidence from auxiliary targets.

\begin{table}[tb]
\caption{Learning targets supplied by each type of execution evidence. Completed sampled paths provide policy returns; checked transition and continuation fields provide auxiliary supervision.}
\label{tab:evidence-routing}
\centering\small
\setlength{\tabcolsep}{4pt}
\begin{tabularx}{\linewidth}{@{}>{\raggedright\arraybackslash}p{2.7cm}Y Y@{}}
\toprule
Evidence & Policy component & Auxiliary component \\
\midrule
Completed actor path & Replayed success, endpoint distance, invalid-attempt and action costs. & Verified transition fields may supply targets. \\
Applicability and effects & Inapplicable attempts count toward $I_j$; no positive no-op reward. & Predict the attempted action's actual effects or select a valid alternative. \\
Searched repair/rejoin & No success label or sampled-policy return is assigned from the witness. & Imitate a replay-checked repair, or a repair with a checked suffix. \\
Collector interruption & Exclude the incomplete comparison; do not assign failure or success. & Independently verified local fields remain usable. \\
Task-horizon stop & Use the observed terminal label, including noncompletion. & Retain available transition targets. \\
Shared or virtual record & Preserve the declared tree-slot weights and original physical identity. & Deduplicate targets while retaining ancestry. \\
\bottomrule
\end{tabularx}
\end{table}

\paragraph{Conditional credit and normalization.}
At the root fork, the two values are the means of the three leaf scores under each prefix. At a nonterminal suffix fork, the three values are its individual complete-path scores. Prefix terms are constant within a suffix group and cancel on centering. Each fork uses population standard deviation and $\tau=10^{-4}$. If all advantages in a root tree are zero, that tree is skipped; otherwise its observed forks retain equal weight, including a zero-signal fork if present.

For a root tree $r$, write $\mathcal F_r$ for its observed forks, $\mathcal A_f$ for the branches at fork $f$, and $N_{f,a}$ for response tokens in the corresponding physical segment. The implemented reduction is
\begin{equation}
\mathcal L_r=\frac{1}{|\mathcal F_r|}\sum_{f\in\mathcal F_r}
\frac{1}{|\mathcal A_f|}\sum_{a\in\mathcal A_f}
\frac{1}{N_{f,a}}\sum_{i=1}^{N_{f,a}}
\left[\ell_{\mathrm{clip}}(\rho_i,\widetilde A_{f,a})+\beta k_i\right],
\label{eq:implemented-loss}
\end{equation}
where $\ell_{\mathrm{clip}}$ is the negative clipped term in Eq.~\ref{eq:tree-surrogate}. Each eligible root receives one optimizer update per epoch in seeded shuffled order. Prompt tokens have no loss. The leaf-credit control instead averages complete-path token losses over the six weighted leaves. In that control, a shared prefix appears in several loss terms by design; it remains one physical collection segment.

Computational reuse preserves the statistical weights of the original samples. A shared decision is scored once, and each path that uses it contributes its loss with the original coefficient. We count collected decisions, tokens processed during backpropagation, and repeated appearances in the loss separately. This accounting separates computational reuse from a change in the learning objective.

\paragraph{Frozen reference and behavior likelihoods.}
Before optimization, the trainer caches reference response-token log probabilities from the round's starting checkpoint. It retains the actual generation probabilities in the denominator of $\rho_i$ and checks their numerical agreement with the unmodified model. Left padding, position IDs, and the original physical generation batches are preserved during rescoring. The regularizer uses
\begin{equation}
\Delta_i=\log p_{\mathrm{ref}}(y_i\mid x_i)-\log p_\theta(y_i\mid x_i),
\qquad k_i=\exp(\Delta_i)-1-\Delta_i.
\label{eq:sampled-kl}
\end{equation}
Under current-policy sampling, the expectation of $k_i$ equals the forward KL to the reference. During optimization, the tokens remain those sampled by the collection policy, so $k_i$ acts as a sampled regularizer. The reference is the starting checkpoint of each round. The certificate-outcome implementation uses clipping $\epsilon=0.2$, $\beta=0.04$, and a fixed final-epoch checkpoint.

\paragraph{Aligned supervision in the joint update.}
For the full joint objective, the outcome role selects an available, verified target and its field mask. Cross-entropy is applied only at the selected target positions. An invalid attempted action can be an input to consequence prediction; it is not a positive action target. Candidate-selection targets identify a verified preferred candidate, and repair targets require replayable actions. Empty pools are removed before normalizing mixture proportions. A retained example's identity records whether it enters the mixture or the separate Union-retention channel; intentional multiple uses retain their total weight.

The full objective combines this policy update with the selected Union and retention targets. The current 1.5B policy stage sets $\lambda_{\mathrm{mix}}=\lambda_{\mathrm{Union}}=0$, following its supervised initialization. An auxiliary control can change the selected targets while preserving the policy component. Collection reuse is controlled separately because sharing computation need not alter either the examples or their weights. These distinctions make it possible to test learning from additional outcomes and reducing the cost of obtaining them as separate contributions.

%% file: sections/14_final_benchmark_details.tex
\section{Benchmark Configuration and Reproduction}
\label{app:final-benchmark}
\label{app:evaluation}

\subsection{Tasks and Execution}

The five-model comparison contains 99 Main tasks, 51 Long tasks, 100 Recovery cases, and 100 Deviation cases. Every model uses the same identifiers within each panel. Main includes cutting/staging, cleaning/staging, disposal/restoration, and retrieval/placement, with 25, 25, 24, and 25 tasks. Long combines procedures with executable reference lengths of six to twelve actions. A reference supplies a valid completion route and its length; it need not be a shortest path.

Recovery starts at saved states following an intervention. Deviation balances 50 invalid-feedback cases and 50 valid deviations. An invalid action preserves the world and supplies error feedback; a valid deviation changes the state before the model continues. Reference prefixes and interventions are replayed to reconstruct each start. Every model then receives the same 12-decision allowance. Task and case sets can overlap across panels, so results are reported per panel rather than pooled into a grand total.

The planner observes the goal, current location, visible entities and relations, recent actions, and execution feedback. The EgoVine input adapter deterministically formats these permitted fields for the model, omitting the separate belief-graph serialization. Predictions never overwrite the environment state or observation memory. Success is the executor's goal check within the decision cap. Parsing rejection consumes one decision; a parsed action that fails a precondition returns execution feedback and also consumes one decision.

Main and Long request 12 candidates at each decision. Each parsed candidate is checked on a cloned state. Selection orders candidates by applicability, then smaller successor goal distance, then goal satisfaction, with the first proposal breaking ties. The selected action executes in the live world. Recovery and Deviation request and execute one candidate. All candidates, including unselected and malformed responses, contribute to proposal counts. This common selector supplies the same execution-based ranking rule to all five models.

For EgoVine, 12-candidate decoding uses temperature 0.2, top-$p$ 1, top-$k$ 0, and a 256-token response limit. Single-candidate recovery decoding is greedy with the same limit. No constrained JSON grammar or action repair is applied. The results use the fixed first inference replicate. The saved task, case, protocol, and model identities bind each result to its execution record.

\subsection{Evaluated Checkpoints}

Both EgoVine models use Qwen2.5-Instruct bases with rank-16 LoRA adapters, alpha 32, and zero adapter dropout on attention and MLP projections. The fixed adapters after supervision and policy learning are identified below. The 1.5B model is the repair-v2 checkpoint, loaded as one cumulative adapter. These weights are used throughout the current benchmark and the paired analyses in Appendix~\ref{app:v2-behavior}.

\begin{table}[htb]
\centering\small
\begin{tabularx}{\linewidth}{@{}lY@{}}
\toprule
Model & Adapter SHA256 \\
\midrule
EgoVine-0.5B & \texttt{f2d5ad5dc677c16e971f8af41a7c9437}\newline\texttt{76d0d9cf7230f8c88c2ad1e991146e69} \\
EgoVine-1.5B & \texttt{96f9c8f83cd646c7e36704016354a62f}\newline\texttt{ba336dc4d148d4eeb96eaa49acdc14b5} \\
\bottomrule
\end{tabularx}
\caption{Fixed checkpoint identities for all current EgoVine results.}
\label{tab:v2-checkpoints}
\end{table}

The external model names identify the recorded services: \texttt{qwen-flash} through DashScope's OpenAI-compatible interface, \texttt{deepseek-flash} through the DeepSeek API, and \texttt{gpt-5.6-luna} through GitHub Copilot CLI. Task definitions, permitted observations, parsing, candidate allocation, action selection, and goal checks are shared. Each adapter formats the permitted fields for its model. Service identifiers specify the evaluated endpoints; the local EgoVine decoding parameters above describe the two local models.

The benchmark evaluates the resulting planners. Appendix~\ref{app:final-training} reports comparisons with their exact supervised parents. Appendix~\ref{app:completed-settings} retains the earlier 0.5B training controls with their original initialization and objectives.

\subsection{Training Recipes for the Evaluated Checkpoints}
\label{app:current-recipes}

\paragraph{Union initialization for 0.5B.}
\label{app:union-initialization}
The Union dataset contains 9,502 serialized training rows across 14 modes on ten videos, with 179 diagnostic rows on three reused development videos. Multiple target views can share one physical context. The initial pool includes 14 unique direct-recovery rows, repeatedly sampled before the later recovery expansion. Historical consequence and route labels retain source hashes and cross-view checks; not every source transition was independently replayed during this initial construction. Negative rejoin labels denote bounded-search outcomes.

Initialization starts from fresh LoRA on Qwen2.5-0.5B-Instruct, with seed 2026092501, rank 16, alpha 32, no dropout, a BF16 base, and FP32 adapters. Each update draws 16 core and two recovery examples, with field and mode weights. Three epochs perform 1,779 committed updates at learning rate $3\times10^{-5}$. The fixed final-epoch adapter begins with SHA256 \texttt{89e1e6d994a747e9}. This checkpoint starts both the current recovery-expanded lineage and the earlier controls in Appendix~\ref{app:05b-development}.

\paragraph{0.5B training lineage.}
From this Union initialization, recovery-expanded supervised training uses 14,402 rows for one epoch and 1,186 updates. Each update draws eight core and five recovery examples, with recovery loss mass 0.25, learning rate $10^{-5}$, and seed 2026092701. The resulting supervised parent has adapter hash beginning \texttt{1087f65899f82a73}. It supplies both the starting and reference policy for a fresh collection from 48 roots. The final native conditional argument-fork update runs one epoch with 28 informative optimizer steps, learning rate $2\times10^{-6}$, clipping 0.1, restricted KL coefficient 0.1, full KL coefficient 0.01, and seed 2026092601. This policy stage uses no additional retention examples or branch-preference pairs. The final adapter is the 0.5B checkpoint in Table~\ref{tab:v2-checkpoints}.

\paragraph{1.5B training lineage.}
The supervised parent starts with fresh LoRA on Qwen2.5-1.5B-Instruct and learns 360 executor-verified reference-action examples from 47 tasks on ten training videos. Eight complete epochs give 360 optimizer updates at batch size eight and learning rate $2\times10^{-5}$, with seed 2026092417. The fixed eighth-epoch adapter, beginning \texttt{4f9a7b24c2748a69}, supplies the collection and reference policy. This supervised stage uses reference-action targets.

The subsequent collection contains 94 roots: 47 normal contexts and 47 manipulation-error contexts. Each root uses two prefixes of at most three actions and three continuations per prefix, with a 24-action horizon. The recorded likelihood check admits 89 roots (47 normal and 42 error contexts); five error groups fail this numerical check and are excluded before optimization. Task-horizon failures remain eligible observed outcomes. Conditional full-response updates follow Appendix~\ref{app:implementation-binding}, with learning rate $2\times10^{-6}$, clipping 0.2, KL coefficient 0.04, and seed 2026092513. One epoch performs 38 informative updates and processes 72,169 response tokens during backpropagation; 51 zero-variance groups produce no update. Auxiliary replay is disabled during this policy stage. The fixed final adapter is the current repair-v2 checkpoint.

These recipes realize the framework through successive supervision and policy stages. The 0.5B checkpoint uses Union and recovery-expanded targets; the 1.5B checkpoint tests reference-action supervision followed by full-response branch learning. These recipes identify the supervised targets and policy updates behind each fixed checkpoint.

The accompanying \path{benchmark_v2/training_controls} records parent and final hashes, training protocols, evaluation identities, and per-episode outcomes. 

\subsection{Aggregation, Uncertainty, and Work}
\label{app:benchmark-uncertainty}

Table~\ref{tab:benchmark-system-r0} and Figure~\ref{fig:benchmark-overview} are generated from the same per-task records. Appendix~\ref{app:v2-behavior} expands the two-checkpoint analysis to 100 paired tasks or cases per panel, retaining the same deployment protocol. Five-model comparisons and two-model analyses each state their own denominators.

Paired success differences are computed before aggregation. Percentile 95\% intervals use 20,000 task-cluster bootstrap draws, retaining related cases within each sampled task; video-cluster intervals provide a sensitivity analysis. These intervals describe variation over the evaluated instances for the fixed checkpoints. They are unadjusted for multiple comparisons and do not sample new training runs.

\input{tables/benchmark_v2_external_intervals}

Table~\ref{tab:v2-external-intervals} reports all comparisons between the two EgoVine checkpoints and the three external models. Main differences relative to Qwen Flash and DeepSeek Flash remain positive under both cluster choices. Differences from Luna are less precisely resolved: the observed Main, Recovery, and Deviation leads have intervals spanning zero, and the 1.5B Long result ties Luna. Figure highlights identify the observed highest values within a panel.

\input{tables/benchmark_v2_decisions}

A decision is one model--environment turn. Main and Long generate 12 candidates per turn, and the intervention panels generate one. Table~\ref{tab:v2-decisions} includes all successes and failures. Candidate counts measure generated proposals, while model requests, batching, and environment checks are separate operations. The paper compares decisions and candidates without equating service latency with local model compute. Training collection, verification, and optimization belong to a separate cost account.

\paragraph{Portable reproduction.}
The full V2 experiment bundle includes the selected task records, summary tables, paired statistics, and compressed execution traces. The portable paper source includes the frozen plotted records and derived execution diagnostics in \path{benchmark_v2}, together with the current training-control records; raw compressed trajectories remain in the full experiment bundle. The offline script \path{scripts/plot_benchmark_v2.py} rebuilds every current benchmark figure and table. The trace-analysis script recomputes behavioral counts from the full V2 package without new inference.

%% file: tables/benchmark_v2_external_intervals.tex
\begin{table}[H]
\centering\small
\setlength{\tabcolsep}{4pt}
\begin{tabularx}{\linewidth}{@{}Yrrr@{}}
\toprule
Comparison & Difference & Task-cluster 95\% CI & Video-cluster 95\% CI \\
\midrule
\multicolumn{4}{@{}l}{\textit{Main ($N=99$)}} \\
0.5B $-$ Qwen Flash & +35.4 & $[22.2, 47.5]$ & $[17.2, 52.5]$ \\
0.5B $-$ DeepSeek Flash & +20.2 & $[8.1, 31.3]$ & $[4.2, 36.7]$ \\
0.5B $-$ Luna & +5.1 & $[-6.1, 16.2]$ & $[-10.3, 20.2]$ \\
1.5B $-$ Qwen Flash & +37.4 & $[24.2, 50.5]$ & $[20.0, 54.5]$ \\
1.5B $-$ DeepSeek Flash & +22.2 & $[9.1, 34.3]$ & $[6.6, 38.7]$ \\
1.5B $-$ Luna & +7.1 & $[-5.1, 19.2]$ & $[-7.4, 22.2]$ \\
\addlinespace
\multicolumn{4}{@{}l}{\textit{Long ($N=51$)}} \\
0.5B $-$ Qwen Flash & +29.4 & $[15.7, 43.1]$ & $[11.3, 52.2]$ \\
0.5B $-$ DeepSeek Flash & +11.8 & $[-3.9, 27.5]$ & $[-10.9, 39.2]$ \\
0.5B $-$ Luna & -7.8 & $[-23.5, 7.8]$ & $[-26.5, 12.1]$ \\
1.5B $-$ Qwen Flash & +37.3 & $[19.6, 54.9]$ & $[16.7, 55.4]$ \\
1.5B $-$ DeepSeek Flash & +19.6 & $[2.0, 37.3]$ & $[1.8, 38.6]$ \\
1.5B $-$ Luna & +0.0 & $[-15.7, 15.7]$ & $[-11.7, 11.1]$ \\
\addlinespace
\multicolumn{4}{@{}l}{\textit{Recovery ($N=100$)}} \\
0.5B $-$ Qwen Flash & +30.0 & $[14.4, 45.1]$ & $[9.8, 48.9]$ \\
0.5B $-$ DeepSeek Flash & +15.0 & $[-3.0, 32.7]$ & $[-7.6, 36.4]$ \\
0.5B $-$ Luna & +7.0 & $[-10.0, 24.5]$ & $[-12.1, 24.3]$ \\
1.5B $-$ Qwen Flash & +20.0 & $[4.0, 36.0]$ & $[0.0, 38.8]$ \\
1.5B $-$ DeepSeek Flash & +5.0 & $[-11.0, 21.2]$ & $[-14.1, 24.8]$ \\
1.5B $-$ Luna & -3.0 & $[-18.4, 13.0]$ & $[-17.5, 11.2]$ \\
\addlinespace
\multicolumn{4}{@{}l}{\textit{Deviation ($N=100$)}} \\
0.5B $-$ Qwen Flash & +33.0 & $[18.0, 47.5]$ & $[13.6, 51.4]$ \\
0.5B $-$ DeepSeek Flash & +17.0 & $[-1.0, 34.7]$ & $[-6.2, 38.3]$ \\
0.5B $-$ Luna & +10.0 & $[-7.0, 27.0]$ & $[-8.7, 26.9]$ \\
1.5B $-$ Qwen Flash & +22.0 & $[6.0, 37.6]$ & $[2.9, 40.4]$ \\
1.5B $-$ DeepSeek Flash & +6.0 & $[-10.2, 22.5]$ & $[-13.4, 25.8]$ \\
1.5B $-$ Luna & -1.0 & $[-16.3, 14.9]$ & $[-15.3, 13.0]$ \\
\bottomrule
\end{tabularx}
\caption{Paired success differences in percentage points on the five-model common task sets. Each row subtracts the external model from the named EgoVine checkpoint. Both interval columns use 20,000 cluster-bootstrap draws of the same paired outcomes; task and video clustering retain their respective groups. All six comparisons are shown for each panel.}
\label{tab:v2-external-intervals}
\end{table}

%% file: tables/benchmark_v2_decisions.tex
\begin{table}[H]
\centering\small
\begin{tabularx}{\linewidth}{@{}Yrrrr@{}}
\toprule
Model & Main & Long & Recovery & Deviation \\
\midrule
Qwen Flash & 8.52 & 10.53 & 8.52 & 8.66 \\
DeepSeek Flash & 7.41 & 9.25 & 7.45 & 7.50 \\
Luna & 6.47 & 8.37 & 6.76 & 6.91 \\
EgoVine 0.5B & 6.51 & 9.39 & 6.67 & 6.55 \\
EgoVine 1.5B & 6.41 & 8.55 & 7.87 & 7.79 \\
\bottomrule
\end{tabularx}
\caption{Mean decisions in the five-model comparison, including failed episodes. Candidate counts are 12 times these values for Main and Long and equal to them for Recovery and Deviation.}
\label{tab:v2-decisions}
\end{table}

%% file: sections/17_v2_behavior.tex
\section{Paired Checkpoint Behavior}
\label{app:v2-behavior}

\subsection{Task Outcomes and Execution Cost}

\begin{figure}[tb]
\centering
\includegraphics[width=\linewidth]{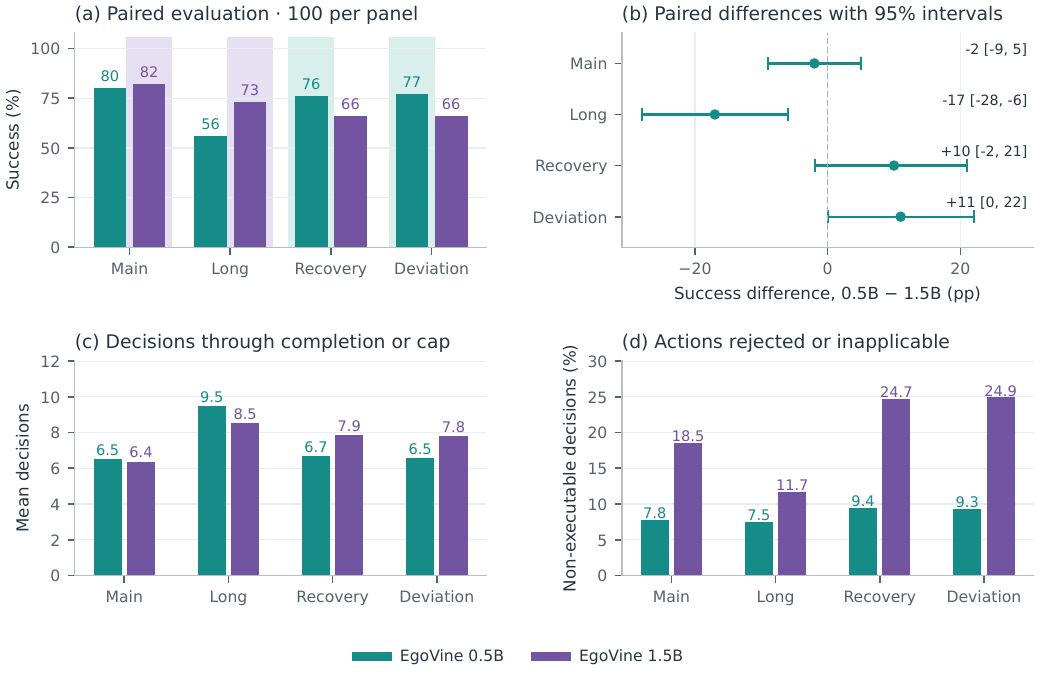}
\caption{Two-checkpoint comparison on 100 paired instances per panel. (a) Success. (b) Paired success differences and task-cluster 95\% intervals. (c) Mean decisions over all instances. (d) Fraction of decisions yielding a parser rejection or inapplicable action. Pale backgrounds mark the higher success rate within this two-model comparison.}
\label{fig:v2-behavior}
\end{figure}

Figure~\ref{fig:v2-behavior} compares the fixed EgoVine checkpoints on 100 identical tasks or cases per panel. Their Main success rates are 80\% and 82\%, and their Long rates are 56\% and 73\%, for 0.5B and 1.5B respectively. Recovery and Deviation favor 0.5B in point estimate. The task-cluster intervals in the figure use paired differences oriented as 0.5B minus 1.5B. Video-cluster sensitivity intervals are $[-12.4,7.5]$ on Main, $[-28.2,-4.4]$ on Long, $[-3.8,22.7]$ on Recovery, and $[-2.7,23.6]$ on Deviation. The Long difference remains negative under both cluster choices. Recovery differences are less precisely resolved across sources.

\input{tables/benchmark_v2_execution}

The distinction between parsed and executable actions is material. On Recovery, 1.5B has 34 parsing rejections and 160 parsed but inapplicable attempts in 787 decisions; 0.5B has no parsing rejections and 63 inapplicable attempts in 667 decisions. Among parsed actions, the inapplicable fractions remain 160/753 (21.2\%) and 63/667 (9.4\%), respectively. Recovery failures therefore involve action prerequisites as well as response format. On Long, 1.5B has a higher non-executable fraction but reaches more goals, making subgoal selection relevant alongside local validity.

\begin{figure}[tb]
\centering
\includegraphics[width=\linewidth]{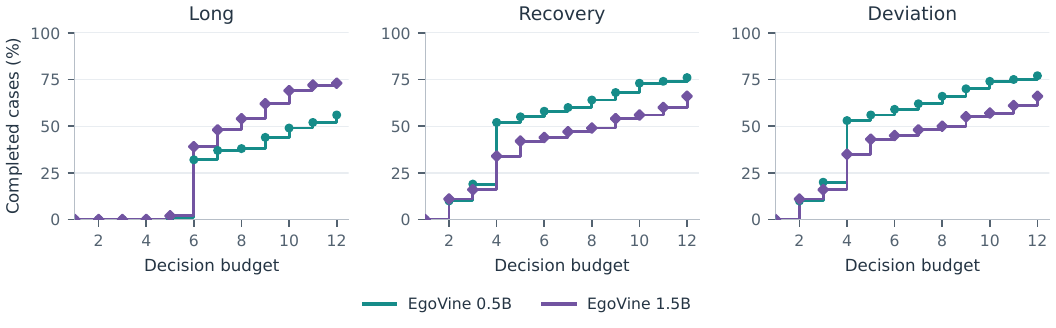}
\caption{Completion at prefixes of the same recorded 12-decision trajectories, over 100 paired cases per panel. Curves show how quickly cases finish; they do not involve new rollouts for each budget. All panels retain the shared benchmark deployment protocol; intervention cases execute one action per turn.}
\label{fig:v2-budget}
\end{figure}

\begin{figure}[tb]
\centering
\includegraphics[width=\linewidth]{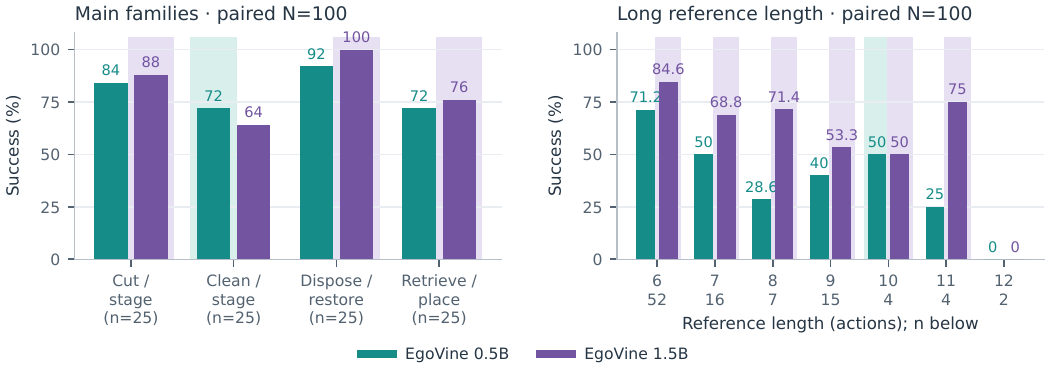}
\caption{Two-model family and reference-length profiles on the complete 100-task panels. Counts below the axis specify each group. These strata belong to the paired two-model analysis; Figure~\ref{fig:benchmark-overview} uses the five-model task sets.}
\label{fig:v2-scale-groups}
\end{figure}

The budget curves agree with the aggregate profile: 1.5B completes more Long tasks earlier, whereas 0.5B completes more intervention cases early. At decision four, Recovery completion is 52 versus 34 and Deviation completion is 53 versus 35. The mean differences include the entire evaluated panel, including episodes that use their full allowance without success.

\FloatBarrier
\subsection{Preparation and Temporary Regression}
\label{app:v2-preparation-analysis}

Let $d_t$ count the unsatisfied required predicates and task milestones after decision $t$, each with unit weight. An applicable action is counted as preparation when it changes the world and $d_t=d_{t-1}$, and as regression when $d_t>d_{t-1}$. Navigation, retrieving a required object, and opening a container can prepare later success without immediately reducing goal distance. A valid action that leaves the world unchanged is counted separately. Table~\ref{tab:v2-preparation} counts successful episodes containing at least one action of each kind.

\input{tables/benchmark_v2_preparation}

All successful Long episodes contain preparation, and many contain temporary regression. Excluding navigation leaves preparation in all 56 and 73 successful episodes. Excluding both navigation and pickup leaves 29/56 and 39/73. Preparation also occurs in failed episodes: 37/44 and 18/27 after excluding navigation. This pattern describes the task's demand for enabling actions; their value depends on the continuation. The bottle example in Appendix~\ref{app:valid-alternatives} checks two feasible continuations and compares their costs. These execution measures describe behavior; prediction accuracy would compare the model's consequence fields with executed effects.

\subsection{Paired Trajectory Examples}
\label{app:v2-cases}

\paragraph{Choosing the next subgoal.}
The Long case from video \texttt{P01-20240203-130505}, composite task \texttt{899bc474377f399d}, requires cutting a carrot and placing it on a board, then washing, drying, and staging the board. The reference length is nine. Both checkpoints perform the first three actions successfully, reaching goal distance three. Table~\ref{tab:v2-board-case} shows their subsequent behavior. Every action in both traces is executable.

\begin{table}[H]
\centering\small
\begin{tabularx}{\linewidth}{@{}lYY@{}}
\toprule
Decisions & EgoVine-0.5B & EgoVine-1.5B \\
\midrule
1--3 & Pick up, cut, and place carrot; $d:7\to3$. & Pick up, cut, and place carrot; $d:7\to3$. \\
4--6 & Visit sink, board, then sink; $d=3$. & Pick up board, visit sink, wash board; $d:3\to2$. \\
7--9 & Wash and dry sink, visit board; $d=3$. & Dry board, visit destination, place board; $d:2\to0$. \\
10--12 & Pick up and replace carrot, visit sink; $d:3\to5\to3$. & Task already complete after decision 9. \\
\bottomrule
\end{tabularx}
\caption{A paired Long case where local applicability does not resolve the next subgoal. $d$ denotes goal distance.}
\label{tab:v2-board-case}
\end{table}

\paragraph{Retrieval before closure.}
The Recovery case from video \texttt{P01-20240202-175627} asks the actor to retrieve \texttt{drawer\_02} from \texttt{fridge\_01}, close the storage, and place the object on \texttt{dishwasher\_03}. The case identifier ends in \texttt{recovery::forced\_valid\_deviation}. From the saved post-intervention state, 0.5B retrieves the object, closes the fridge, navigates, and places it, completing in four decisions. Its first action leaves goal distance at two while establishing the holding precondition for placement.

The 1.5B model first closes the fridge, reducing goal distance from two to one, and navigates to the destination. Placement and pickup attempts fail. It reopens the fridge at decision ten, retrieves the object at eleven, and closes at twelve; placement remains unfinished. This example shows the tension between immediate predicate progress and preserving the prerequisites for the remaining plan. The two cases illustrate the measured behavioral profiles and retain their original outcomes.

\FloatBarrier
\subsection{Additional Native-Interface Comparison}

A separate evaluation uses the same fixed checkpoints on 169 paired cases from 61 tasks: 61 normal starts, 61 invalid-feedback cases, and 47 valid deviations. It allows 24 decisions, executes one sampled action per turn, and uses temperature 1, top-$p$ 1, and top-$k$ 0 with the common evaluation seed. Its task construction and horizon are kept separate from the 12-decision benchmark.

\begin{figure}[tb]
\centering
\includegraphics[width=\linewidth]{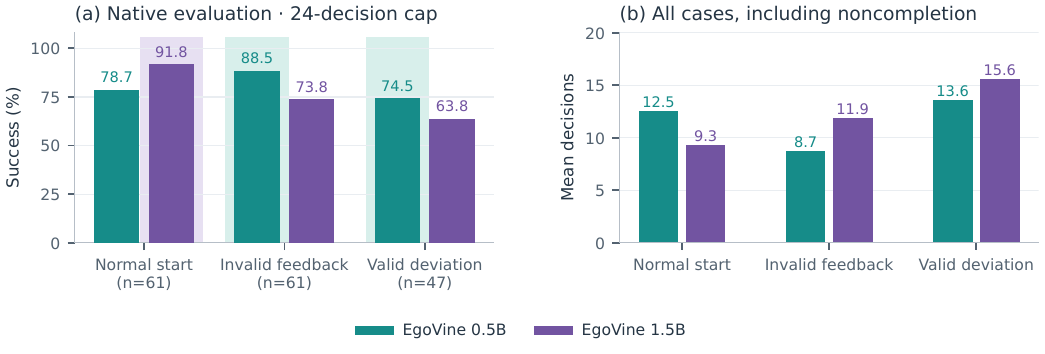}
\caption{Additional native-interface comparison of the same two checkpoints. Left: success by starting condition. Right: mean decisions including failed cases. All 169 cases use a 24-decision cap and one sampled action per turn.}
\label{fig:v2-native}
\end{figure}

The 0.5B model completes 137/169 cases and 1.5B completes 131/169. Normal starts favor 1.5B, at 56/61 versus 48/61; invalid-feedback and valid-deviation cases favor 0.5B, at 54/61 versus 45/61 and 35/47 versus 30/47. Figure~\ref{fig:v2-native} shows the corresponding decision counts. The paired task-macro difference is $+2.46$ points for 0.5B, with a 95\% task-cluster interval of $[-4.92,9.84]$. This additional evaluation exhibits the same planning--recovery contrast under its own task and decoding conditions. Appendix~\ref{app:final-training} compares the 1.5B model with its exact parent over both inference seeds; Appendix~\ref{app:k1} examines single-candidate deployment on fixed 12-decision subsets.

%% file: tables/benchmark_v2_execution.tex
\begin{table}[H]
\centering\small
\setlength{\tabcolsep}{4pt}
\begin{tabularx}{\linewidth}{@{}Yrrrrr@{}}
\toprule
Panel / model & Successes & Decisions & Rejected & Inapplicable & No change \\
\midrule
Main / 0.5B & 80 & 654 & 0 & 51 & 91 \\
Main / 1.5B & 82 & 638 & 11 & 107 & 73 \\
\addlinespace
Long / 0.5B & 56 & 947 & 0 & 71 & 78 \\
Long / 1.5B & 73 & 854 & 13 & 87 & 60 \\
\addlinespace
Recovery / 0.5B & 76 & 667 & 0 & 63 & 38 \\
Recovery / 1.5B & 66 & 787 & 34 & 160 & 69 \\
\addlinespace
Deviation / 0.5B & 77 & 655 & 0 & 61 & 34 \\
Deviation / 1.5B & 66 & 779 & 44 & 150 & 69 \\
\bottomrule
\end{tabularx}
\caption{Execution counts on 100 paired cases per panel. Rejected responses yield no parsed action. Inapplicable actions are parsed but fail execution. No-change actions pass applicability but leave the world unchanged; this is distinct from preparation that changes the world without reducing goal distance.}
\label{tab:v2-execution}
\end{table}

%% file: tables/benchmark_v2_preparation.tex
\begin{table}[H]
\centering\small
\begin{tabularx}{\linewidth}{@{}Yrrrr@{}}
\toprule
 & \multicolumn{2}{c}{With preparation} & \multicolumn{2}{c}{With temporary regression} \\
Panel & 0.5B & 1.5B & 0.5B & 1.5B \\
\midrule
Main & 73/80 & 74/82 & 14/80 & 16/82 \\
Long & 56/56 & 73/73 & 24/56 & 28/73 \\
Recovery & 68/76 & 59/66 & 7/76 & 11/66 \\
Deviation & 69/77 & 59/66 & 7/77 & 11/66 \\
\bottomrule
\end{tabularx}
\caption{Successful episodes containing at least one preparatory or regressive action. Preparation changes the world while keeping goal distance unchanged; regression increases goal distance. Each denominator is the number of successful episodes for that model and panel.}
\label{tab:v2-preparation}
\end{table}

%% file: sections/18_current_training_controls.tex
\section{Training-Stage Comparisons for the Current Checkpoints}
\label{app:final-training}

These comparisons follow the exact adapters evaluated in the current benchmark back to their supervised parents. Each parent/final comparison holds the tasks, visible observations, parsing, decoding, and decision allowance fixed. All stages use 24 decisions with one directly executed action per turn, temperature 1, top-$p$ 1, and top-$k$ 0. The retained episode records, parent identities, and paired statistics accompany \path{benchmark_v2/training_controls}.

\subsection{0.5B: Expanded Supervision and the Subsequent Policy Update}

The current 0.5B lineage begins with the Union SFT checkpoint specified in Appendix~\ref{app:union-initialization}. It then receives expanded recovery supervision, followed by a conditional policy update. The same 61 tasks, 169 cases, and two evaluation seeds yield 338 episodes for each stage. Table~\ref{tab:current-05b-stages} reports all three stages on this fixed panel.

\begin{table}[H]
\centering\small
\setlength{\tabcolsep}{4pt}
\begin{tabularx}{\linewidth}{@{}Yrrrr@{}}
\toprule
Condition & Episodes & Union SFT & Recovery SFT & Final EgoVine \\
\midrule
All cases & 338 & 215 & 269 & 274 \\
Initial planning & 122 & 83 & 97 & 101 \\
Invalid feedback & 122 & 83 & 106 & 106 \\
Valid deviation & 94 & 49 & 66 & 67 \\
\midrule
Training-source cases & 260 & 180 & 227 & 235 \\
Reused-development cases & 78 & 35 & 42 & 39 \\
\bottomrule
\end{tabularx}
\caption{Success counts along the current 0.5B training lineage. Recovery SFT is the exact parent of the final adapter in Table~\ref{tab:v2-checkpoints}. Source slices partition the same 338 episodes; they are not additional evaluations.}
\label{tab:current-05b-stages}
\end{table}

Expanded supervised training raises success from 215/338 to 269/338. The paired task-macro improvement is $15.03$ percentage points, with a 95\% task-cluster interval of $[8.47,22.27]$. It improves both initial planning and intervention recovery. The subsequent policy update reaches 274/338: four additional initial-planning successes and one additional valid-deviation success, with unchanged invalid-feedback success. Its task-macro improvement over the exact parent is $1.37$ points, with interval $[-1.78,4.51]$. These intervals use 4,000 paired task-cluster resamples, retaining both evaluation seeds and related cases together.

The supervised stage accounts for most of the observed improvement and adds both recovery experience and training exposure. The policy update gives a smaller positive aggregate change, while the reused-development count falls from 42 to 39. Stage-wise reporting places each change at the point in training where it occurs.

Across the complete panel, Recovery SFT uses 3,878 decisions and the final checkpoint uses 3,834; invalid-action rates are 15.52\% and 15.13\%. Recorded generation time is 1,738.83 versus 1,791.99 seconds on the same local setup. Fewer decisions therefore do not imply lower measured wall time. These costs describe the resulting closed-loop trajectories; a throughput comparison would hold the workload fixed.

\subsection{1.5B: Planning and Recovery from a Shared Supervised Parent}

Both 1.5B checkpoints are evaluated on the same 169 cases from 61 tasks under both inference seeds, using the shared settings above. The reference-action SFT parent completes 131 and 130 cases; the final checkpoint completes 131 and 132. Table~\ref{tab:supp-native} reports both repeats and their pooled episode counts. The overall change is 261/338 to 263/338, or $+0.59$ percentage points, with a paired task-cluster 95\% interval of $[-2.31,3.33]$ and video-cluster interval $[-2.94,3.83]$. Both inference repeats belong to the same training lineage.

\input{tables/supp_native_parent}

The source-specific task-macro differences are $+0.35$ points on training-source cases and $+1.19$ on reused-development cases, with task-cluster intervals $[-2.48,3.19]$ and $[-4.76,7.14]$. These task-macro statistics weight parent tasks equally within each source stratum; the overall count above weights episodes equally. Each of the 338 case--seed pairs is evaluated by both checkpoints, giving 676 model--episode records.

\paragraph{Targeted recovery panel.}

The current 1.5B checkpoint starts from a fresh reference-action SFT parent and then learns from branches collected at normal and perturbed states, as specified in Appendix~\ref{app:current-recipes}. A matched recovery evaluation contains 60 cases from 49 tasks: interventions either block the reference continuation or undo goal progress. Both models receive the same actor-visible observations and evaluation seed. The panel uses existing training and development worlds; 42 of its saved error contexts also occur among the training roots.

\begin{table}[H]
\centering\small
\setlength{\tabcolsep}{4pt}
\begin{tabularx}{\linewidth}{@{}Yrrrr@{}}
\toprule
Checkpoint & Successes & Task macro (\%) & Distance reduction & Decisions \\
\midrule
Reference-action SFT & 32/60 & 51.02 & 0.543 & 16.64 \\
Final EgoVine & 35/60 & 56.12 & 0.618 & 16.22 \\
\bottomrule
\end{tabularx}
\caption{Matched 1.5B recovery comparison. Distance reduction and decisions are task-macro means, including failed cases. Final EgoVine uses the same adapter as all current 1.5B benchmark results.}
\label{tab:current-15b-stages}
\end{table}

The policy update completes three additional cases and improves mean goal-distance reduction (Table~\ref{tab:current-15b-stages}). The paired task-macro success difference is $5.10$ points, with a 95\% task-cluster interval of $[-4.08,15.31]$. The positive recovery estimate has a wide interval. The reduction in remaining goal distance describes progress within these intervention cases, complementing the completion counts in the full panel.

Both size-specific comparisons follow one training lineage. They connect the final benchmark adapters to observed changes during training, while the earlier auxiliary and credit controls below retain their original parents and objectives.

%% file: tables/supp_native_parent.tex
\begin{table}[htb]
\centering\small
\setlength{\tabcolsep}{4pt}
\begin{tabularx}{\linewidth}{@{}Yrrr@{}}
\toprule
Inference seed & 1.5B SFT parent & Final EgoVine & Difference (pp) \\
\midrule
2026092461 & 131/169 & 131/169 & +0.00 \\
2026092462 & 130/169 & 132/169 & +1.18 \\
\midrule
Both repeats & 261/338 & 263/338 & +0.59 \\
\bottomrule
\end{tabularx}
\caption{Exact-parent comparison on 169 native cases, repeated with two inference seeds. All rows use the same fixed pair of checkpoints; the final row pools episodes from both repeats.}
\label{tab:supp-native}
\end{table}

%% file: sections/13_completed_experiments.tex
\section{Prior Development Controls}
\label{app:completed-settings}
\label{app:05b-development}

Earlier development controls used the same Union initialization to examine auxiliary supervision and policy credit before the recovery-expanded lineage. They are retained in the accompanying experiment archive with their original objectives, checkpoints, and execution records. They are not used as headline evidence for the current benchmark; the released checkpoints and their active recipes are specified in Appendix~\ref{app:final-training}.

The current paper therefore keeps the development lineage for provenance while focusing the appendix results on the fixed checkpoints, their paired behavior, direct-actor subset, and measured computational reuse.

%% file: sections/20_supplementary_experiments.tex
\section{Learning, Prediction, and Computational Reuse}
\label{app:supplementary}

\subsection{Protocols and Experimental Units}

The follow-up studies connect the framework's learning targets, policy updates, and reuse operations to controlled measurements. They preserve the benchmark checkpoints in Table~\ref{tab:v2-checkpoints}. Newly trained branches are separate experimental conditions, evaluated at their prescribed final update. The broader 1.5B parent comparison is reported in Appendix~\ref{app:final-training}.

The learning controls use 114 saved starts from 14 parent tasks on three previously used development videos. These comprise 14 full initial states, 42 invalid-feedback starts, 30 valid deviations, and 28 reference-suffix starts. The seven-task Long group is a subset of the 14 initial tasks. Evaluation directly executes one greedy action per turn, with 12 decisions and a 72-token output limit. Each paired control uses the same saved starts, permitted observations, parser, and execution rules.

Seeds 2026092611 and 2026092612, abbreviated 11 and 12 below, repeat continued training from the same parent. For the paired learning controls, we average the two differences within each parent task and use 5,000 task-cluster bootstrap draws for percentile 95\% intervals. The source package includes checkpoint identities, training protocols, and derived statistics in \path{supplement_20260926}; the full experimental bundle additionally contains raw trajectories and replay programs.

\subsection{Supplementary Training Controls}
\label{app:followup-learning}

The supplementary controls use 114 saved starts from 14 parent tasks and preserve the same parser, observations, decision allowance, and checkpoint identities. They probe how collected experience can be replayed under supervised and policy updates, how repair targets enter the mixture, and how collector refresh changes the realized branches. These short-budget controls are diagnostic: the headline benchmark and final checkpoint recipes do not rely on a claim that a particular policy-update variant is uniformly superior. Seeds, realized workloads, and derived statistics remain in the accompanying experiment archive for reproducibility.

\subsection{Auxiliary Target Diagnostics}
\label{app:prediction}

The paper's endpoint is completed task behavior under the shared executable-action benchmark. Auxiliary probes of consequence fields, candidate-related outputs, and bounded rejoin labels are therefore treated as diagnostic checks on the supervision interface rather than as standalone capability claims. They are not used to rank models or to attribute the benchmark scores. The complete probe records and scoring scripts remain in the accompanying experiment archive.

\subsection{Direct-Actor Deployment on a Fixed Subset}
\label{app:k1}

We select 24 Main and 24 Long starts by a fixed family, reference-length, and hash rule, then evaluate each exact parent/final pair at $K=1$. The visible input projection, parser, temperature 0.2, 256-token response limit, and 12-decision allowance are shared. Each proposed action executes directly. Three starts occur in both panels: four checkpoints therefore yield 180 actual episodes and 192 panel memberships. Most starts continue an existing procedure (21 Main and 19 Long), so these scores describe the selected task states.

\begin{figure}[htb]
\centering
\includegraphics[width=\linewidth]{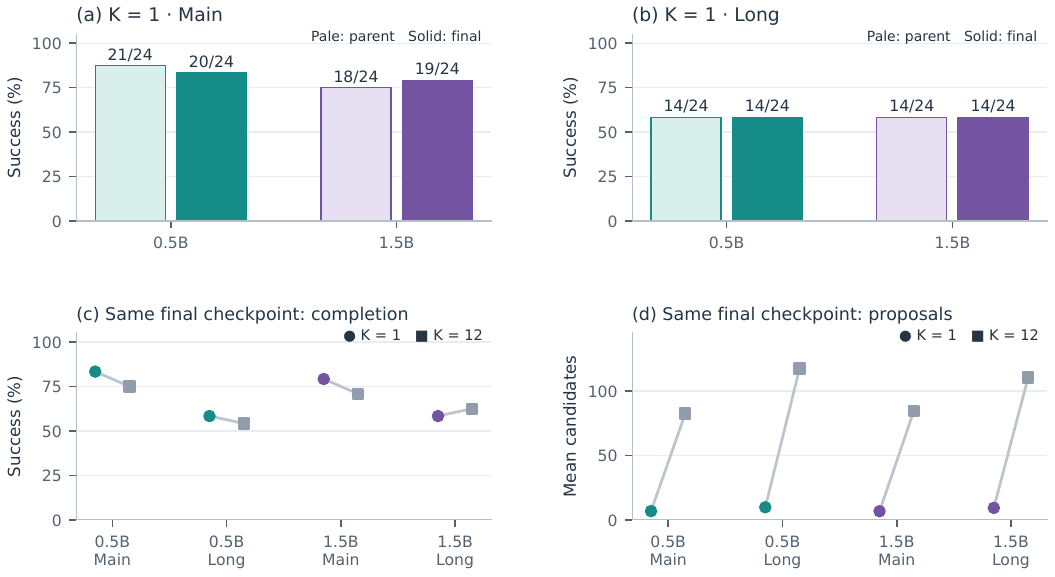}
\caption{Direct-actor deployment on the fixed 24-start panels. (a,b) Exact supervised parents and final checkpoints. (c,d) The same final weights under direct deployment and the retained benchmark deployment, showing completion and generated proposals. All decisions, including failures, contribute to the recorded work. Parents and final checkpoints share the evaluation settings within each panel.}
\label{fig:supp-k1}
\end{figure}

\input{tables/supp_k1}

Under direct deployment, the final actors complete 20/24 and 19/24 Main starts and 14/24 Long starts each. These directly executed decisions give similar completion to their supervised parents. Table~\ref{tab:supp-k1} also reports the same final weights under the retained benchmark deployment.

\FloatBarrier
\subsection{Fixed-Workload Computational Reuse}
\label{app:reuse-measurements}

\begin{figure}[htb]
\centering
\includegraphics[width=\linewidth]{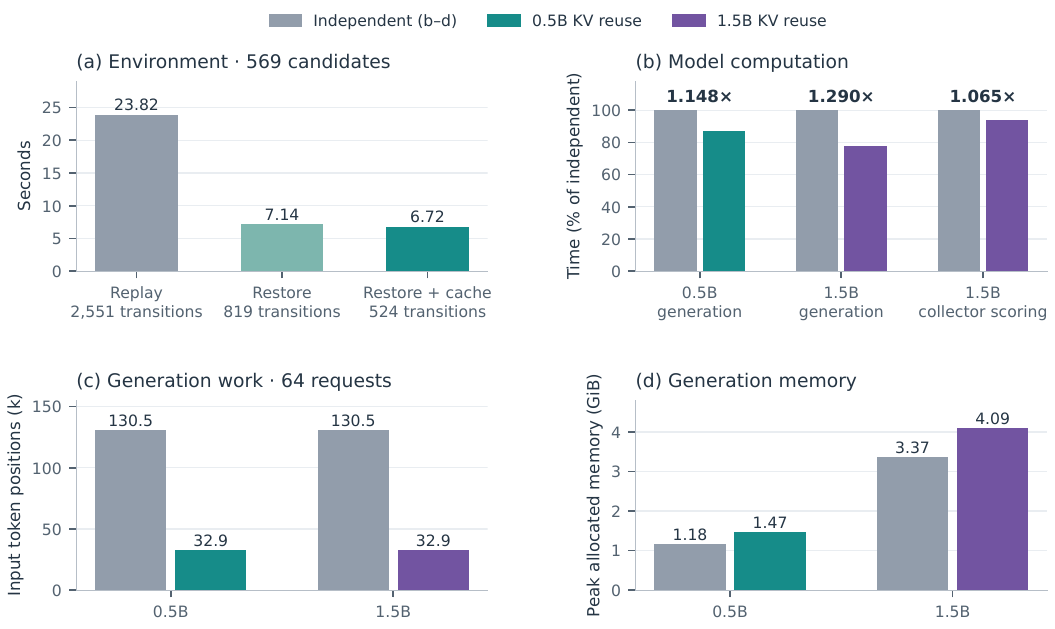}
\caption{Compatible reuse under fixed workloads. (a) CPU execution time and physical transition counts. (b) Model time normalized to each workload's independent mode; labels give speedup. (c,d) Input token positions and peak allocated memory for 64-request generation. Generation uses five repeats per mode; collector scoring uses three. Timings include cache operations. Each workload is a separate comparison.}
\label{fig:supp-reuse}
\end{figure}

BunchFusion separates the reuse of deterministic transitions from the reuse of model computation. We measure these operations with the work and outputs fixed within each comparison. Figure~\ref{fig:supp-reuse} shows their time, work, and memory profiles, and Table~\ref{tab:supp-efficiency} gives absolute values.

\paragraph{Environment execution.}
The CPU workload contains 569 candidates from 72 states. Independent prefix replay, snapshot restoration, and restoration with exact transition caching produce the same checked execution evidence. Across five repeats, median time falls from 23.820 to 7.140 to 6.720 seconds; physical transitions fall from 2,551 to 819 to 524. Timings include resetting, snapshot creation, cache construction and lookup, and verification. The final configuration reuses 295 transitions after restoration has removed repeated prefix execution.

\paragraph{Generation and scoring.}
The generation workload has 64 requests sharing 16 exact prefixes, with 48 cache hits. Supplied verb prefixes, request-local random generators, and batch size one preserve every generated token across modes. After one full-workload warmup, five repeats yield speedups of $1.148\times$ for 0.5B and $1.290\times$ for 1.5B. Forward calls fall from 445 to 397 and from 474 to 426, respectively. Cache residency increases peak allocated memory, as shown in the figure. Each comparison runs within its model's host; absolute times across sizes are not a hardware-matched scaling study.

A further workload uses 248 retained requests from 24 deterministically selected roots of the actual 1.5B collector, loaded under its original collection weights. Three repeats reduce median fixed-token scoring time from 141.138 to 132.542 seconds ($1.065\times$). The modes score the same response tokens with the same masks and leaf weights; their maximum log-probability difference is zero in the measured runtime. Timings include cache construction, copying, lookup, and CUDA synchronization, excluding model loading and disk export.

\input{tables/supp_efficiency}
\FloatBarrier

The timings quantify environment execution, generation, and scoring on their measured workloads. End-to-end collection and training, including suffix/rejoin reuse, require a full-cycle measurement. The full 94-root collection ledger records 5,047 physical decisions, 6,125 logical path references, and 564 logical leaves. Logical references preserve sharing and are not additional samples. In that ledger, strict request-level compatibility leaves fewer reuse opportunities than matching world state alone, which motivates measuring each operation on its actual compatible workload.

\FloatBarrier
\subsection{Refreshing the Collector}
\label{app:refresh}

For each continuation seed, we start a second update stage from its joint-update actor and collect from the same 24 training roots using either the original supervised collector or the updated actor. Root query/probe caps are 32/8, with two draws per eligible sibling. Both arms then use 48 updates and 384 auxiliary draws. Behavior probabilities come from the actual collector; the KL reference remains the original supervised parent in both arms. This instantiates collector refresh with a fixed reference, whereas Algorithm~\ref{alg:egovine} describes the general round-wise reference schedule.

\input{tables/supp_refresh}

Refreshing changes the collected branches and targets: the two fixed-collector pools have 35 and 37 groups, while the refreshed pools have 32 and 33. Both stage-1 actors complete 61/114 evaluation cases. After the second stage, fixed collectors yield 58 and 60 successes, and refreshed collectors yield 59 and 59. The seed-averaged task-macro difference is 0.00 points, with interval $[-2.68,2.68]$. The process completes another collection and update, while providing no consistent performance advantage at this budget. Table~\ref{tab:supp-refresh} records the differing realized work alongside the matched roots and update counts.

%% file: tables/supp_k1.tex
\begin{table}[htb]
\centering\small
\setlength{\tabcolsep}{4pt}
\begin{tabularx}{\linewidth}{@{}llrrrYr@{}}
\toprule
Size & Panel & Parent $K1$ & Final $K1$ & $\Delta$ (pp) & 95\% task CI & Final $K12$ \\
\midrule
0.5B & Main & 21/24 & 20/24 & -4.17 & [-12.50, 0.00] & 18/24 \\
0.5B & Long & 14/24 & 14/24 & +0.00 & [-12.50, 12.50] & 13/24 \\
1.5B & Main & 18/24 & 19/24 & +4.17 & [0.00, 12.50] & 17/24 \\
1.5B & Long & 14/24 & 14/24 & +0.00 & [0.00, 0.00] & 15/24 \\
\bottomrule
\end{tabularx}
\caption{Single-candidate follow-up and retained 12-candidate results on the same fixed subsets. Intervals compare final minus parent at $K=1$. Main and Long overlap on three task starts. The $K=12$ column describes a different deployment configuration for the same final weights.}
\label{tab:supp-k1}
\end{table}

%% file: tables/supp_efficiency.tex
\begin{table}[htb]
\centering\small
\setlength{\tabcolsep}{4pt}
\begin{tabularx}{\linewidth}{@{}Yrrrr@{}}
\toprule
Workload / mode & Seconds & Transitions & Input positions & GiB \\
\midrule
CPU: prefix replay & 23.820 & 2551 & --- & --- \\
CPU: restore & 7.140 & 819 & --- & --- \\
CPU: restore + cache & 6.720 & 524 & --- & --- \\
0.5B generation: independent & 10.683 & --- & 130,485 & 1.177 \\
0.5B generation: KV reuse & 9.305 & --- & 32,907 & 1.475 \\
1.5B generation: independent & 15.430 & --- & 130,514 & 3.372 \\
1.5B generation: KV reuse & 11.960 & --- & 32,936 & 4.092 \\
1.5B scoring: independent & 141.138 & --- & --- & --- \\
1.5B scoring: KV reuse & 132.542 & --- & --- & --- \\
\bottomrule
\end{tabularx}
\caption{Fixed-workload execution and model computation. Times are medians over five repeats for environment and generation, three for scoring. GiB denotes peak allocated GPU memory during generation. Dashes denote metrics outside the reported workload account. Model sizes are timed on their respective hosts.}
\label{tab:supp-efficiency}
\end{table}

%% file: tables/supp_refresh.tex
\begin{table}[htb]
\centering\small
\setlength{\tabcolsep}{4pt}
\begin{tabularx}{\linewidth}{@{}Yrrrrrr@{}}
\toprule
Collector & Groups & Targets & Output tokens & Transitions & Seconds & Success \\
\midrule
Fixed (11) & 35 & 241 & 10,979 & 29,691 & 365.7 & 58/114 \\
Fixed (12) & 37 & 249 & 11,276 & 33,228 & 413.9 & 60/114 \\
Refreshed (11) & 32 & 237 & 10,885 & 36,560 & 433.4 & 59/114 \\
Refreshed (12) & 33 & 239 & 10,756 & 29,620 & 395.5 & 59/114 \\
\bottomrule
\end{tabularx}
\caption{Two-stage collection pilot, with 24 roots, fixed query/probe caps, 48 updates, and 384 auxiliary draws per arm. Both stage-1 actors score 61/114. Actual collection work and branch support vary. Physical transitions include internal reconstruction and clone operations through the executor.}
\label{tab:supp-refresh}
\end{table}